\documentclass{article}

\usepackage[dblblindworkshop, final]{neurips_2025_camera_ready}

\usepackage[utf8]{inputenc} 
\usepackage[T1]{fontenc}    
\usepackage{hyperref}       
\usepackage{url}            
\usepackage{booktabs}       
\usepackage{amsfonts}       
\usepackage{nicefrac}       
\usepackage{microtype}      
\usepackage{xcolor}         

\usepackage{svg}
\usepackage{graphicx}
\usepackage{caption}
\usepackage{subcaption}
\usepackage{float}
\usepackage{amsmath}
\usepackage{multirow}
\usepackage{wrapfig}
\usepackage{tcolorbox}
\usepackage{enumitem}
\usepackage{appendix}
\usepackage{titletoc}

\usepackage{natbib}
\workshoptitle{Efficient Reasoning}

\title{Efficient Reinforcement Learning for Large Language Models with Intrinsic Exploration}

\author{%
  Yan~Sun$^{1,2}$\thanks{Work done during internship at Ant Group}~~~~Jia~Guo$^2$\thanks{Corresponding author}~~~~Stanley~Kok$^1$~~~~Zihao~Wang$^2$~~~~Zujie~Wen$^2$~~~~Zhiqiang~Zhang$^2$ \\
  $^1$National~University~of~Singapore~~~$^2$Ant~Group\\
  \texttt{\{yansun, skok\}@comp.nus.edu.sg}\\
  \texttt{\{jia.g, xiaohao.wzh, zujie.wzj, lingyao.zzq\}@antgroup.com}\\
}

\begin{document}

\maketitle

\begin{abstract}
Reinforcement learning with verifiable rewards (RLVR) has improved the reasoning ability of large language models, yet training remains costly because many rollouts contribute little to optimization, considering the amount of computation required. This study investigates how simply leveraging intrinsic data properties, almost free benefit during training, can improve data efficiency for RLVR. We propose PREPO with two complementary components. First, we adopt prompt perplexity as an indicator of model adaptability in learning, enabling the model to progress from well-understood contexts to more challenging ones. Second, we amplify the discrepancy among the rollouts by differentiating their relative entropy, and prioritize sequences that exhibit a higher degree of exploration. Together, these mechanisms reduce rollout demand while preserving competitive performance. On the Qwen and Llama models, PREPO achieves effective results on mathematical reasoning benchmarks with up to 3 times fewer rollouts than the baselines. Beyond empirical gains, we provide theoretical and in-depth analyses explaining the underlying rationale of our method to improve the data efficiency of RLVR.
\end{abstract}

\section{Introduction}
Reinforcement learning (RL) has become central to improving the reasoning capabilities of large language models (LLMs) by optimizing their self-generated rollouts \citep{guo2025deepseek,team2025kimi}. Recent advances in reinforcement learning with verifiable reward (RLVR) demonstrate that it is a simple yet effective method for scaling reasoning performance \citep{shao2024deepseekmath,yu2025dapo}. However, RLVR still incurs substantial computational overhead, as rollout generation remains the primary training bottleneck \citep{zhong2024optimizing}.

A key source of inefficiency is that not all samples contribute equally to training. On the \emph{prompt side}, some queries are too trivial or too difficult to contribute meaningful gradient \citep{yu2025dapo}. Prompt difficulty is often estimated through pass rates \citep{zhang2025learning} or manually defined criteria \citep{chen2025self,parashar2025curriculum}, but these approaches are expensive and may not reflect the model’s own perception.
On the \emph{rollout side}, responses differ in confidence regardless of correctness (see Fig.~\ref{fig:rollout_entropy}). Low-entropy (confident) responses produce small gradients, whereas highly-entropy (uncertain) responses produce large ones, implying alternative reasoning paths that support exploration (see Appendix~\ref{app:entropy_motivation}). Thus, we may leverage the intrinsic property as an inductive bias for more efficient training.

\begin{figure}[!thb]
 \centering
\includegraphics[width=0.8\linewidth]{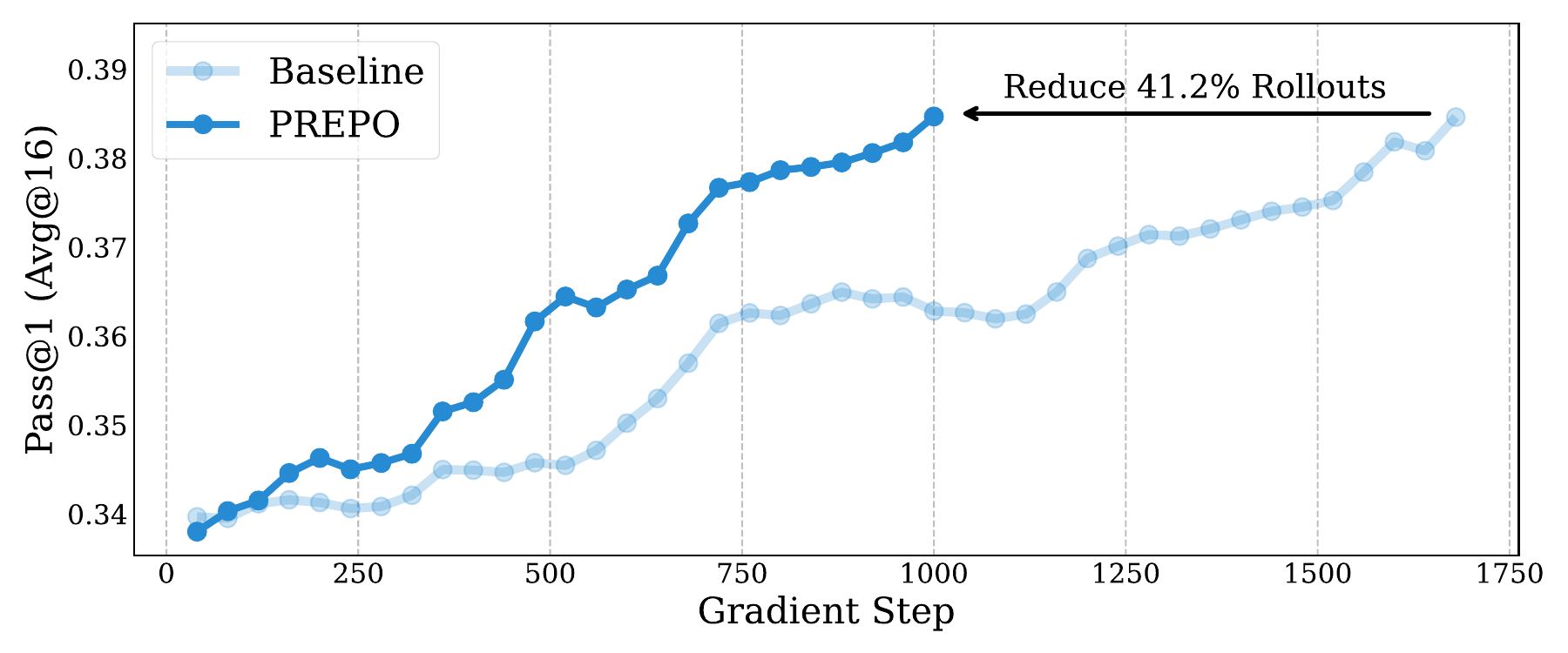}
\caption{Comparison of PREPO and GRPO with random 20\% selection on Qwen2.5-Math-7B, averaged across AIME24, AIME25, MATH-500, and Olympiad Bench.}
\label{fig:main}
\end{figure}

A natural way to improve efficiency is through data selection, i.e., pruning uninformative prompts or rollouts while preserving those that drive learning. There are emerging approaches based on parameterized modeling \citep{qu2025prompt}, replay buffers \citep{liu2025replay}, or selective rollout execution \citep{zheng2025act}. This motivates our research question from a new perspective:

\vspace{.2cm}
\centerline{\textit{Can the intrinsic properties of prompts and rollouts improve the efficiency of RLVR?}}
In this study, we propose \underline{P}erplexity-Schedule with \underline{R}elative-\underline{E}ntropy \underline{P}olicy \underline{O}ptimization (PREPO)\footnote{\href{https://github.com/yan-sun-x/PREPO}{Github Repository: https://github.com/yan-sun-x/PREPO}}, which combines a perplexity-based schedule with sequence-level entropy weighting to realize \textit{intrinsic exploration}. Specifically, PREPO traces perplexity \textit{before} rollout generation to pruning the prompts, and applies entropy weighting \textit{after} rollout generation to emphasize uncertain responses. Across Qwen and Llama models, PREPO surpasses existing data-pruning baselines and remains competitive with the full-data setting, while reducing rollout usage by more than 40\% (see Fig.~\ref{fig:main} for Qwen2.5-Math-7B). These results show that RLVR can be made substantially more efficient by leveraging the intrinsic properties of prompt and rollout data.
\section{Preliminary Analysis}

\begin{wrapfigure}{r}{0.3\textwidth}
\vspace{-1.cm}
\centering
\includegraphics[width=\linewidth]{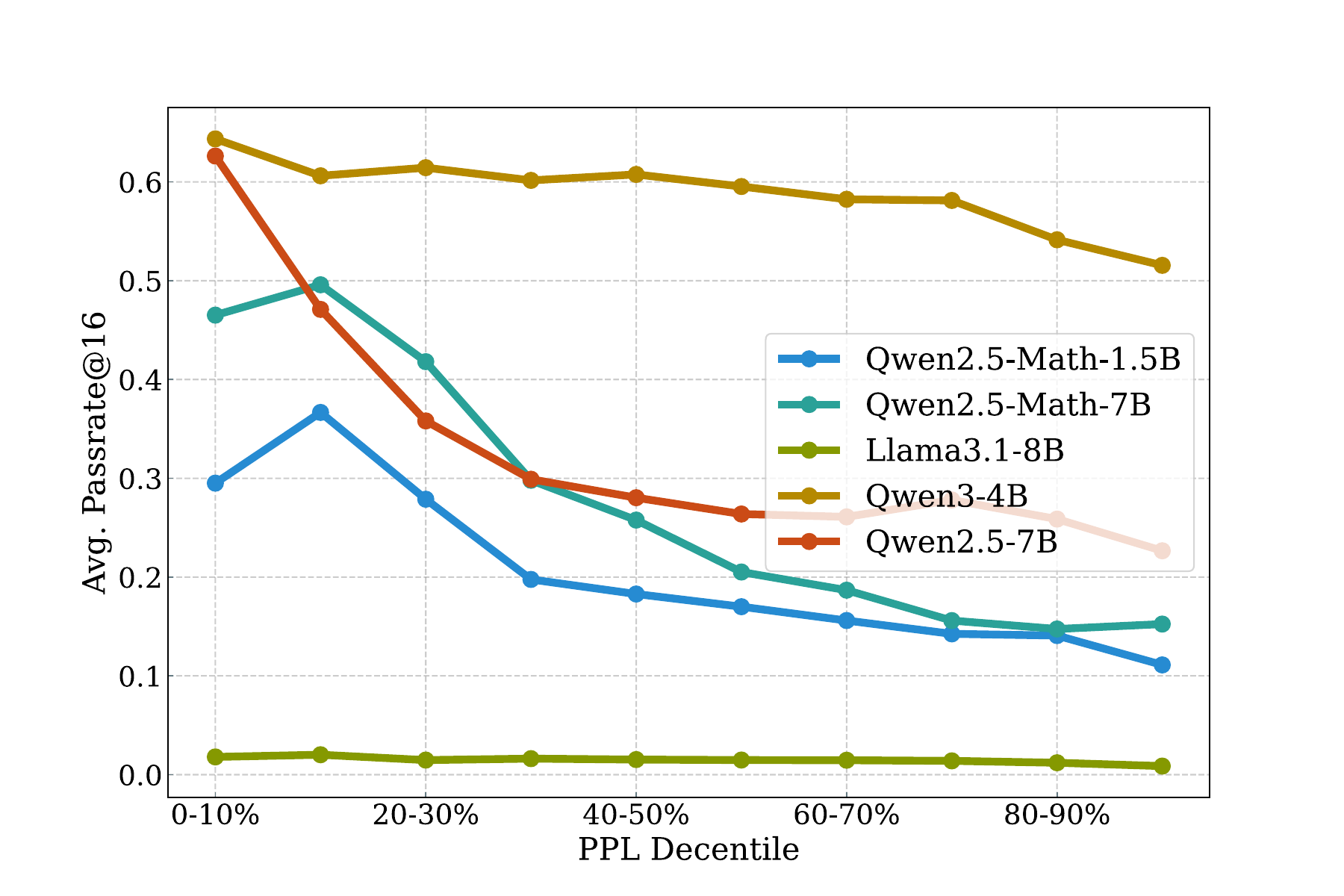}
\caption{Prompt PPL versus average passrate@16.}
\label{fig:ppl-passrate}
\vspace{-.5cm}
\end{wrapfigure}

\subsection{Prompts with Lower-PPL Tend to Yield Higher Passrate}
We begin by examining the relationship between prompt perplexity (PPL) and task difficulty using the DAPO-Math-17K dataset\citep{yu2025dapo}. For both Qwen and Llama models, Figure \ref{fig:ppl-passrate} shows
a clear negative correlation between PPL and passrate@16, where \emph{passrate@16} is the fraction of correct outputs among 16 independent generations for each prompt. Prompts with lower PPL generally yield higher success rates. Table \ref{tab:corr_ppl_passrate} correlation is statistically significant across models, suggesting that PPL can serve as a lightweight signal to identify more informative prompts for training.

\begin{table}[!h]
    \centering
    \caption{Correlation between prompt PPL and \textit{passrate@16}. 
    ($^{***}$ $p < 0.001$, $^{**}$ $p < 0.05$)}
    \label{tab:corr_ppl_passrate}
    \resizebox{\linewidth}{!}{%
    \begin{tabular}{c|c|c|c|c|c|c}
    \toprule
    & Qwen2.5-7B & Qwen2.5-32B & Qwen2.5-Math-7B & Qwen2.5-Math-1.5B & Qwen3-4B & LLama3.1-8B\\\midrule
    Spearman & $-0.233^{***}$ & $-0.207^{***}$ & $-0.183^{**}$  & $-0.186^{***}$ & $-0.169^{***}$ & $-0.199^{***}$ \\
    \bottomrule
    \end{tabular}
    }
\end{table}

\subsection{Training Dynamics of Low-PPL and High-PPL Prompts}

\begin{figure*}[!thb]
    \centering
    \includegraphics[width=\linewidth]{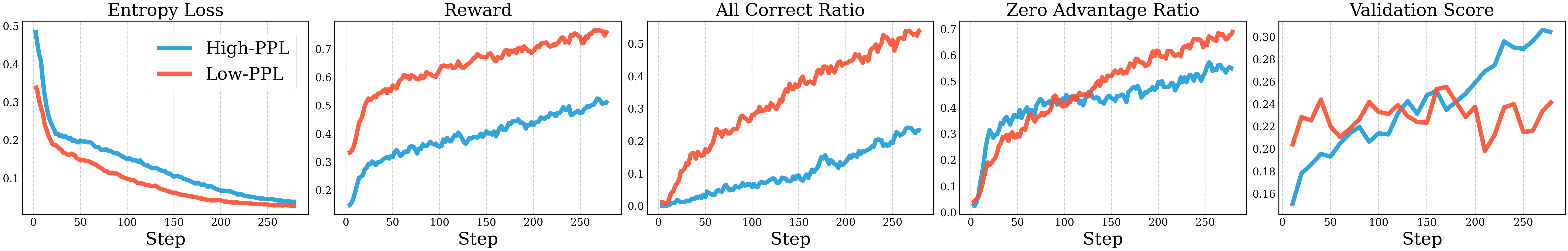}
    \caption{Training dynamics of \textsc{Low-PPL} vs. \textsc{High-PPL} prompts on Qwen2.5-Math-7B.\textit{
    (a) \textsc{High-PPL} prompts have higher entropy.
    (b) \textsc{Low-PPL} prompts have more reward gains.
    (c) \textsc{Low-PPL} prompts reach higher all-correct ratios faster.
    (d) \textsc{Low-PPL} prompts show higher zero-advantage ratios in the later stage.
    (e) \textsc{High-PPL} prompts eventually outperform \textsc{Low-PPL} prompts on AIME24 \citep{aime2024I}.}}
    \label{fig:pilot_study_1}
\end{figure*}

We further compare the training dynamics between the \textsc{Low-PPL} group (i.e., data with the lowest 20\% prompt perplexity) and the \textsc{High-PPL} group (i.e., data with the highest 20\% prompt perplexity) from the entire DAPO-Math-17K dataset.
As illustrated in Figure~\ref{fig:pilot_study_1}, the two groups show distinct behaviors in multiple metrics, where \textsc{Low-PPL} prompts provide stronger learning signals at the early steps, whereas \textsc{High-PPL} prompts retain exploratory benefits that can improve sample efficiency in later steps. Please refer to Appendix \ref{app:low_high_prompt_dynamics} for a detailed analysis of other Qwen and Llama models.

\begin{figure}[!thb]
    \centering
    \includegraphics[width=0.65\linewidth]{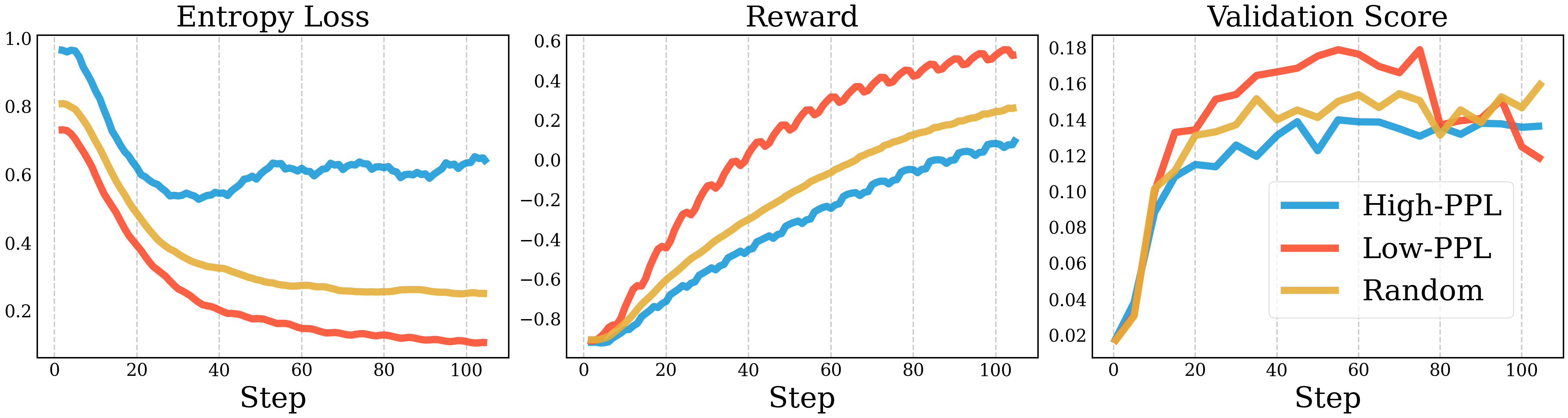}
    \caption{Comparison among \textsc{Low-PPL}, \textsc{High-PPL}, and Random Subsets. \textit{Random lies between the two, showing that PPL-based grouping provides a meaningful pruning signal.}}
    \label{fig:motivation_pilot_random}
\end{figure}

As a baseline, we randomly sample 20\% of the data. As shown in Figure~\ref{fig:motivation_pilot_random}, the \textsc{High-PPL} and \textsc{Low-PPL} groups show distinct behaviors compared to random selection, indicating that PPL-based grouping offers a useful data pruning strategy.

\section{PREPO: \underline{P}PL-Schedule \underline{R}elative-\underline{E}ntropy \underline{P}olicy \underline{O}ptimization}

\subsection{General Online Batch Selection}
Let $\mathcal{B} = \{x_i\}_{i=1}^{|\mathcal{B}|}$ denote the candidate batch at a training step. The goal of online batch selection is to design a mapping
\begin{equation}
\Phi : [0,1] \to 2^{\mathcal{B}}, 
\quad \rho \mapsto \mathcal{I}_\rho,
\end{equation}
where $\rho \in [0,1]$ denotes the normalized training progress, and $\mathcal{I}_\rho \subseteq \mathcal{B}$. The mapping $\Phi$ is required to (i) explicitly depend on $\rho$, so that the distribution of selected samples evolves with training; (ii) the sub-batch size is fixed during training, i.e., $\forall \rho,~|\mathcal{I}_\rho| = K$.

\subsection{PPL-Schedule Online Batch Selection}

For a prompt $x_i = (x_{i,1}, \dots, x_{i,T})$, the perplexity at progress $\rho$ is
\begin{equation}\label{eq:prompt_ppl}
P_i(\rho) = \exp\!\left(-\frac{1}{T}\sum_{t=1}^{T}\log \pi_\rho(x_{i,t}\mid x_i, x_{i,<t})\right),
\end{equation}
where $\pi_\rho$ is the model distribution at progress $\rho$. Since $\pi_\rho$ evolves with training, $P_i(\rho)$ reflects a dynamic difficulty score of the problem. We define the \emph{PPL-schedule} sub-batch as
\begin{equation}
\mathcal{I}_\rho = \{\, \sigma(j) : l(\rho) \le j \le l(\rho)+K-1 \,\},
\end{equation}
where $\sigma$ is the permutation that sorts $\mathcal{B}$ by ascending $P_i(\rho)$. The starting index $l(\rho)$ is given by a linear schedule
\begin{equation}
l(\rho) = \big\lfloor \rho \cdot (|\mathcal{B}| - K) \big\rfloor,
\end{equation}
so that $\mathcal{I}_\rho$ shifts smoothly from prompts with lower PPL to those with higher PPL. While linear scheduling is the simplest case, nonlinear pacing (e.g., quadratic or exponential) can also be used.

\subsection{Relative-Entropy Weighting}
Empirically, we find that training on low-PPL prompts accelerates reward improvement but also leads to a rapid collapse of entropy, thereby reducing exploration. To mitigate this effect during the PPL-schedule, we introduce a sequence-level relative-entropy weighting scheme that adaptively emphasizes uncertain rollouts.

The token-level entropy of a rollout is defined as $H_t = - \sum_{v \in \mathcal{V}} \pi_\theta(v \mid o_{<t}, x)\log \pi_\theta(v \mid o_{<t}, x),$ where $\mathcal{V}$ is the vocabulary. For rollout $i$, the sequence-level entropy is the average across its tokens
\begin{equation}
\bar H_i=\bar H(o_i \mid x) = \frac{1}{|o_i|} \sum_{t=1}^{|o_i|} H_t.
\end{equation} 
The batch-average entropy over $B$ rollouts is
\begin{equation}
\bar H = \frac{1}{B}\sum_{k=1}^{B} \bar H_k .
\end{equation}
The relative weight assigned to rollout $i$ is then given by
\begin{equation}
w_i = \frac{\bar H_i}{\bar H}.
\end{equation}
This formulation ensures that weights are scale-invariant, where rollout’s contribution is determined only by how its entropy compares to the batch mean. Under mild assumptions (see Appendix~\ref{app:theory}), the total weight $\tfrac{1}{B}\sum_i w_i$ remains unbiased with respect to the actual batch size $B$.

Intuitively, this design enables the model to \textit{seek uncertainty within certainty} during the PPL-schedule. While the prompts with lower PPL at early stages typically produce confident (i.e., low-entropy) responses, the weighting mechanism selectively amplifies relatively uncertain (i.e., high-entropy) rollouts, thereby preserving exploratory capacity throughout training.



\subsection{Objective Function}

The PREPO objective integrates PPL-schedule filtering with relative-entropy weighting as below.
\begin{equation}
\resizebox{\linewidth}{!}{$
\begin{aligned}
\mathcal{J}_{\text{PREPO}}(\theta)
&= \mathbb{E}_{\,x \sim \mathcal{I}_\rho,\;\{o_i\}_{i=1}^{G} \sim \pi_{\text{old}}(\cdot \mid x)} \Bigg[
\frac{1}{G}\sum_{i=1}^G w_i \cdot \frac{1}{|o_i|}\sum_{t=1}^{|o_i|}
\min \!\Big(
s_{i,t}(\theta)\,\hat{A}_{i,t},\;
\operatorname{clip}\!\big(s_{i,t}(\theta),\,1-\epsilon_{\text{low}},\,1+\epsilon_{\text{high}}\big)\,\hat{A}_{i,t}
\Big)
\Bigg]
\end{aligned}
$}
\end{equation}
where $\mathcal{I}_\rho$ is the PPL-schedule-filtered batch of prompts at training progress $\rho$, $w_i$ encodes the relative entropy of rollout $i$ at the current micro-batch, $s_{i,t}(\theta)$ is the token-level importance ratio,$ s_{i,t}(\theta) = \frac{\pi_\theta(o_{i,t} \mid x, o_{i,<t})}{\pi_{\text{old}}(o_{i,t} \mid x, o_{i,<t})}$, and $\hat{A}_{i,t}$ is the group-based advantage estimate, $
\hat{A}_{i,t} = \frac{r_i - \operatorname{mean}(\{r_j\}_{j=1}^{G})}{\operatorname{std}(\{r_j\}_{j=1}^{G})}$.

\section{Related Work}
\paragraph{Data efficiency in RLVR.}  
There have been increasing attention to improve data efficiency for RLVR. Several works focus on prompt-side selection. For example, online difficulty filtering \cite{bae2025online}, predictive prompt selection \cite{qu2025prompt}, and curriculum-based methods \cite{zhang2025learning,chen2025self} aim to allocate resources toward more learnable prompts. Other approaches explore importance-based or gradient-informed selection, such as gradient-alignment methods \cite{kamalloo2025learnalign}, influence estimation \cite{chen2025sample}, or policy-advantage based prioritization \cite{wang2025dump}. Parallel efforts reduce redundancy in rollouts, including down-sampling strategies \cite{li2025rollouts} and efficient replay buffer designs \cite{liu2025replay}. These methods highlight the importance of identifying which data truly drives learning.

\paragraph{Entropy Mechanism in RLVR.} 
Entropy has long been studied in reinforcement learning, where entropy-regularized objectives are used to promote exploration in control settings. Recent work extends this perspective to RLVR for reasoning LLMs. \citet{cui2025entropy} show that rapid entropy collapse is a key failure mode, and propose covariance-based updates to slow down decay. \citet{wang2025beyond} further find that high-entropy ``forking tokens'' constitute a small minority of steps yet drive most reasoning improvements, suggesting that entropy signals informativeness at the token level.  



\section{Experiments}\label{para:experiments}

\textbf{Models and Datasets.} We benchmark our method against random selection and GRESO \citep{zheng2025act} across multiple model scales. Specifically, we consider Qwen2.5-7B \citep{team2024qwen2}, Qwen2.5-Math-1.5B, Qwen2.5-Math-7B \citep{yang2024qwen2}, Qwen3-4B (non-thinking) \citep{yang2025qwen3}, and Llama3.1-8B \citep{dubey2024llama}. For training data, we include DAPO \citep{yu2025dapo} and MATH500 datasets \citep{lightman2023lets}.

\textbf{Training and Evaluation.} All models are trained using the \texttt{verl} \citep{sheng2025hybridflow}, with \texttt{vLLM} \citep{Kwon2023EfficientMM} employed for rollout generation to ensure efficient inference. For the Qwen models, evaluation is conducted on a comprehensive suite of benchmarks, including AIME25 \citep{aime2025I}, AIME24 \citep{aime2024I}, MATH500 \citep{lightman2023let}, and OlympiadBench \citep{he2024olympiadbench}, which cover a diverse range of mathematical reasoning challenges. The Llama model is evaluated on MATH500 \citep{lightman2023let} and GSM8K \citep{cobbe2021gsm8k}. We evaluate all models using \textit{pass@1 (avg16)}, i.e., the accuracy of the top-1 response averaged over 16
generations, with temperature 1.

\textbf{Experiment Configuration.} We set the clipping thresholds with $\epsilon_{\text{low}} = 0.2$ by default and a larger $\epsilon_{\text{high}} = 0.28$ for the upper bound. For the Qwen2.5-Math models, we use a maximum context length of 4096 tokens, matching their supported limit. For Qwen3-4B and Llama3.1-8B, we set the context length to 32,768 tokens. Rollouts are generated with temperature $T=1$ using vLLM, producing 8 responses per prompt. The global batch size is 1280, with a reduced batch size fixed at 256 and a mini-batch size of 64. For both PREPO and the baseline, we adopt an online selection ratio of $K/\mathcal{B} = 20\%$ at each training step. For GRESO, we set the targeted
zero-variance percentage as $50\%$. The actor model is optimized with AdamW using a constant learning rate of $1\times 10^{-6}$, momentum parameters $\beta_1 = 0.9$ and $\beta_2 = 0.999$, and a weight decay of 0.01. Following \citet{yu2025dapo}, we omit the KL-divergence regularization term. Training is applied only to the actor parameters and parallelized with Fully Sharded Data Parallel. All experiments are conducted on 32 GPUs.

\subsection{Main Results}
\begin{table}[!h]
\centering
\caption{Performance comparison (\%) on Qwen. \textit{Best results are highlighted in bold or underlined.}}
\label{tab:main_result_qwen}
\begin{tabular}{l|cccc|c|c} 
\toprule
 \textbf{Method}     &      \textbf{AIME25} & \textbf{AIME24} & \textbf{MATH}  & \textbf{Olympiad} & \textbf{Avg $\uparrow$}  & \textbf{\# Rollouts $\downarrow$}  \\ \midrule 
 \textit{Qwen2.5-7B} & 1.25    & 4.17 & 72.26 &  33.09   & 27.69 & -- \\ 
+ Random     &  6.98  & \underline{16.41}  & 75.70 & 38.47 & 34.39 & 716K \\ 
+ GRESO      &  9.22  &  10.83 &  \underline{76.65}  &  \underline{42.07} & 34.59 &  680K \\ 
+ PREPO (Ours)   & \underline{10.21}  & 16.09  & 76.30  & 39.85 & \textbf{35.61} & \textbf{304K} 
\\ \midrule
\textit{Qwen2.5-Math-1.5B} & 3.54   & 10.21  & 55.76 & 27.41  & 24.23 & -- \\ 
+ Random     & \underline{20.00}  & 16.67  & 76.25 & 30.50 & 35.86 & 3.0M \\ 
+ GRESO      &  15.38  &  \underline{20.00} &  \underline{76.65}  &  24.17 & 34.16 & 2.5M \\ 
+ PREPO (Ours)   & \underline{20.00}  & 16.67  & 76.25  & \underline{32.00} & \textbf{36.23} & \textbf{1.1M} \\ \midrule
\textit{Qwen2.5-Math-7B} & 9.17  & 20.80  & 72.26  & 39.56 &  35.45  &  -- \\ 
+ Random              & 10.00  & \underline{26.67}  & 77.80 & \underline{43.26}          & 39.45 & 905K \\ 
+ GRESO      &  18.33  &  25.83 &  77.80  &  26.83 & 37.46 & 654K 
 \\
+ PREPO (Ours)      & \underline{12.81}  & 26.15  & \underline{77.85} & 41.58          & \textbf{39.59} &  \textbf{540K} \\\midrule 
\textit{Qwen3-4B} & 30.00   &  53.33  & 94.10  & 52.67  & 57.53 &  -- \\ 
+ Random      &  60.00  &  70.00 &  96.00  &  59.33 & 71.33 & 553K\\ 
+ GRESO      &  56.67  &  69.17 &  96.40  &  57.33 & 69.89 & 472K \\ 
+ PREPO (Ours)    & \underline{66.67}  & \underline{80.00} & \underline{96.60}  & \underline{60.67}  & \textbf{75.99}  &  \textbf{348K}\\ 
\bottomrule
\end{tabular}
\end{table}



\begin{table}[!htb]
\centering
\caption{Performance comparison (\%) on Llama. \textit{Best results are highlighted in bold or underlined.}}
\label{tab:main_result_llama}
\begin{tabular}{l|cc|c|c} 
\toprule
 \textbf{Method}     & \textbf{GSM8K} & \textbf{MATH} & \textbf{Avg $\uparrow$} & \textbf{\# Rollouts $\downarrow$} \\ 
\midrule
\textit{Llama3.1-8B} &  9.53 & 6.05  &  7.79 & --\\ 
+ Random             & 46.63 & 14.60 & 30.61 & 266K \\ 
+ GRESO              & 41.77 & 16.80 & 29.29 & 273K \\ 
+ PREPO (Ours)       & \underline{51.10} & \underline{21.81} & \textbf{36.55} & \textbf{115K} \\ 
\bottomrule
\end{tabular}
\end{table}


\textbf{PREPO achieves up to $3\times$ (66.7\%) rollout reduction while matching or surpassing baseline performance across all models.}
As shown in Table \ref{tab:main_result_qwen} and \ref{tab:main_result_llama}, PREPO reduces rollout amount by $3\times$ (63.3\%) on Qwen2.5-Math-1.5B, $1.7\times$ (40.3\%) on Qwen2.5-Math-7B, $1.6\times$ (37.1\%) on Qwen3-4B (non-thinknig), $2.4\times$ (57.5\%) on Qwen2.5-7B, and $2\times$ (48.9\%) on Llama3.1-8B. These results demonstrate that PREPO consistently improves data efficiency, often reducing rollout demand by two to three times, without sacrificing, and in many cases even improving, benchmark performance. More analyses, including ablations, are provided in Appendix \ref{app:additional_experiments}.

Due to space constraints, additional discussion and analysis of PREPO are provided in Appendix~\ref{app:discussion} and Appendix~\ref{app:theory}.

\subsection{Ablation Study}

We conduct an ablation study to isolate the effect of relative-entropy weighting on top of the PPL-schedule. We report the performance of PPL-schedule at the same number of rollouts as PREPO. As shown in Table~\ref{tab:ablation}, PREPO consistently outperforms the PPL-schedule across most benchmarks, indicating that entropy-based rollout weighting provides additional gains beyond prompt-side scheduling.

\begin{table}[h]
\centering
\caption{Performance comparison (\%) between PPL-chedule and PREPO. \textit{Best results are highlighted in bold or underlined.}}
\label{tab:ablation}
\resizebox{\linewidth}{!}{%
\begin{tabular}{l|c|cccc|c}
\toprule
Model & Method & AIME25 & AIME24 & MATH & Olympiad Bench & Avg $\uparrow$\\
\midrule
\multirow{2}{*}{Qwen2.5-Math-7B} 
& PPL-schedule &  10.00 & 23.33 & 74.60 & 39.21 & 36.79 \\ 
& + Relative-entropy (PREPO) & \underline{12.81} & \underline{26.15} & \underline{77.80} & \underline{41.58} & \textbf{39.59} \\
\midrule
\multirow{2}{*}{Qwen2.5-7B} 
& PPL-schedule &  6.98 & \underline{16.41} & 75.70 & 38.47 & 34.39 \\
& + Relative-entropy (PREPO) & \underline{10.20} & 16.09 & \underline{76.30} & \underline{39.85} & \textbf{35.61}\\
\bottomrule
\end{tabular}
}
\end{table}

\section{Discussion}\label{app:discussion}

\begin{figure}[!h]
    \centering
    \begin{subfigure}[t]{0.45\textwidth}
        \centering
        \includegraphics[width=\linewidth]{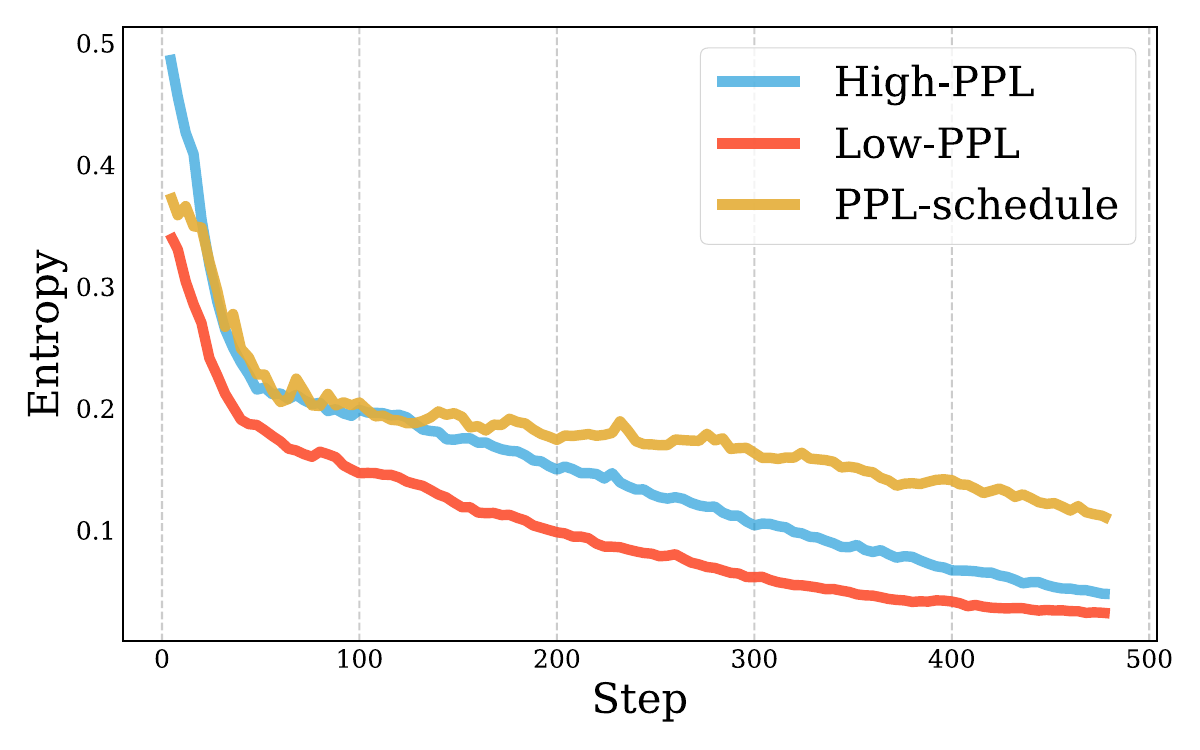}
        \caption{Entropy}
    \end{subfigure}%
    ~ 
    \begin{subfigure}[t]{0.45\textwidth}
        \centering
        \includegraphics[width=\linewidth]{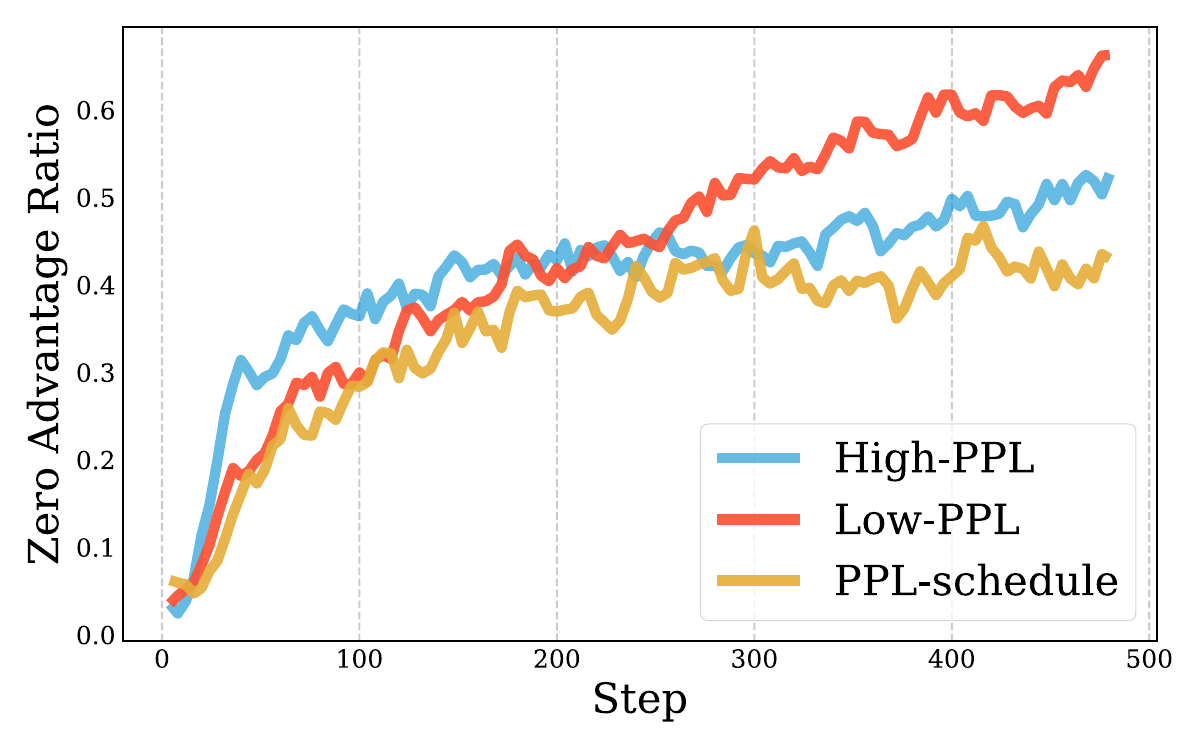}
        \caption{Zero Advantage Ratio} 
    \end{subfigure}
    \caption{Comparison of training dynamics between PPL-schedule and static PPL selection (Low- and High-PPL groups)}
    \label{fig:discussion_ppl_curriculum}
\end{figure}

\paragraph{What Does the PPL-Schedule Contribute to Training?}
PPL-schedule helps maintain an adequate degree of exploration throughout learning. As illustrated in Figure~\ref{fig:discussion_ppl_curriculum}, we compare three training configurations for Qwen2.5-Math-7B: (1) training exclusively with high-PPL prompts, (2) exclusively with low-PPL prompts, and (3) a PPL-schedule that gradually transitions from low- to high-PPL prompts. The PPL-schedule yields a more balanced optimization trajectory than either static regime. In terms of entropy loss, PPL-schedule shows a slower decline, avoiding the rapid entropy collapse that typically occurs under low-PPL-only training. Regarding the zero-advantage ratio, the PPL-schedule consistently achieves lower values, meaning that a larger proportion of rollouts provide non-trivial gradient contributions.

\begin{figure}[!htb]
    \centering
    \begin{subfigure}[t]{0.45\textwidth}
        \centering
        \includegraphics[width=\linewidth]{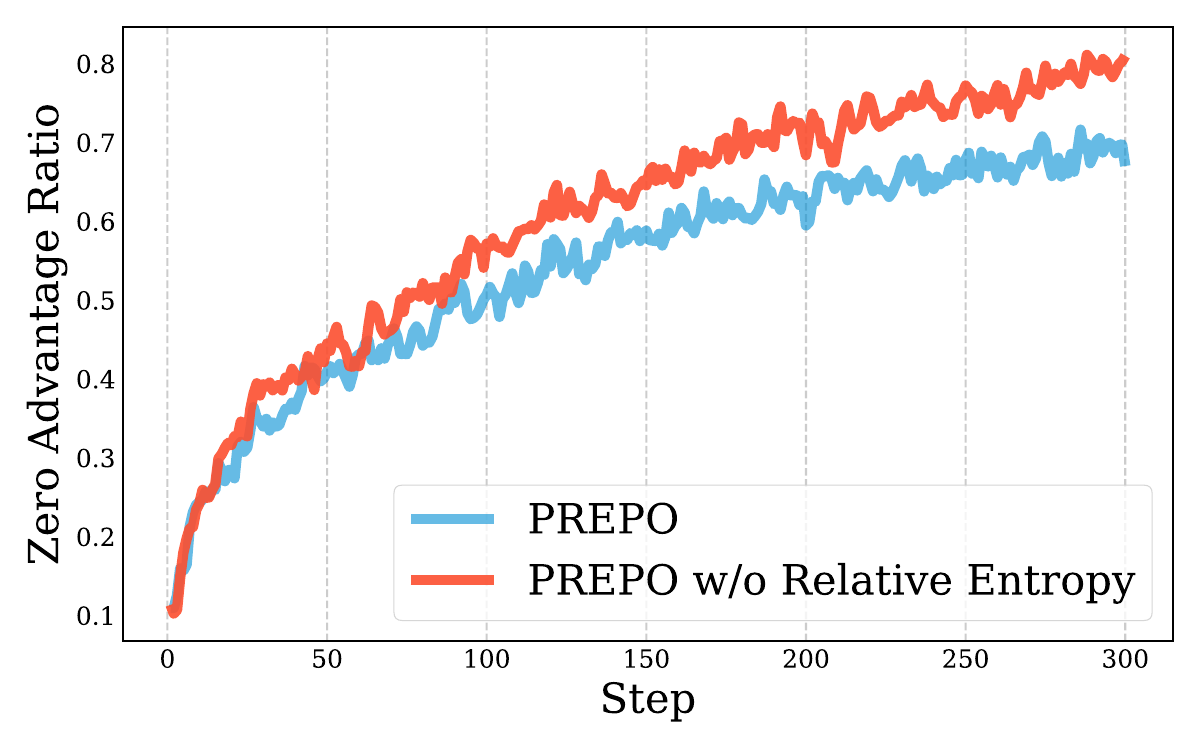}
        \caption{Qwen2.5-Math-7B}
    \end{subfigure}%
    ~ 
    \begin{subfigure}[t]{0.45\textwidth}
        \centering
        \includegraphics[width=\linewidth]{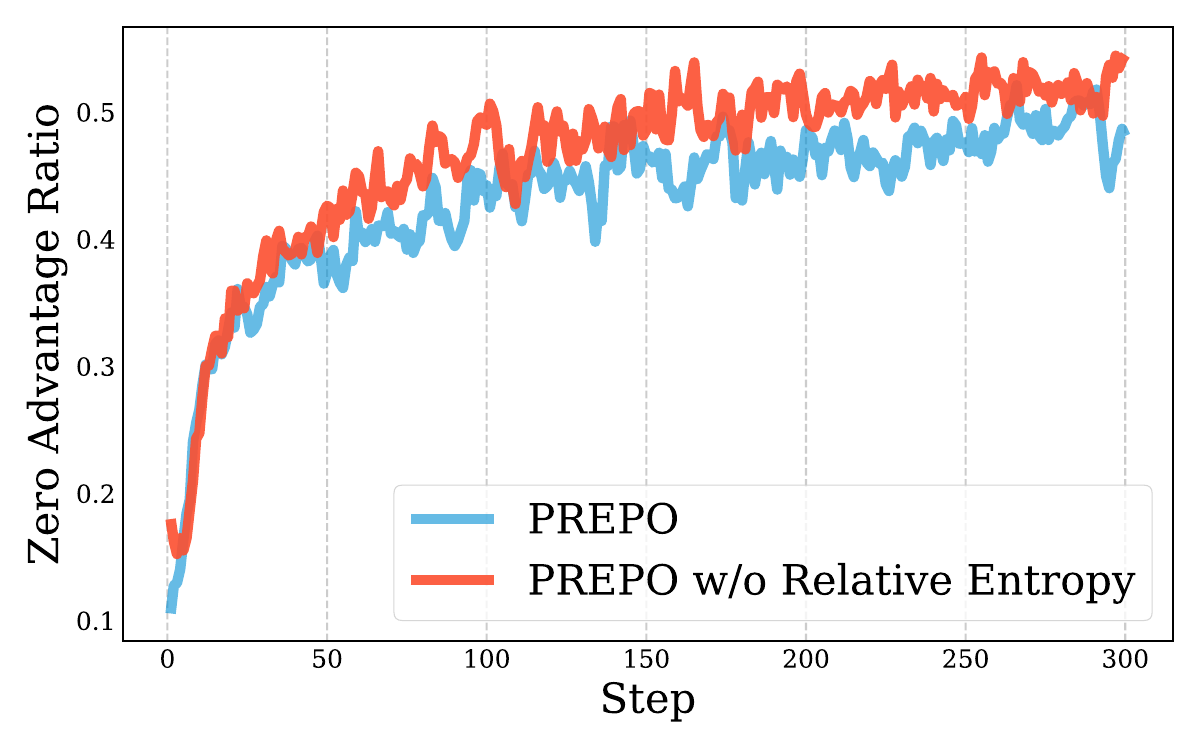}
        \caption{Qwen2.5-Math-1.5B} 
    \end{subfigure}
    \caption{Comparison of zero advantage ratio between PPL-schedule and PREPO.}
    \label{fig:discussion_relative_entropy}
\end{figure}

\paragraph{What does relative-entropy bring to the training?}
Introducing relative entropy into PREPO further enhances training efficiency. As shown in Figure~\ref{fig:discussion_relative_entropy}, incorporating relative-entropy weighting further reduces the zero-advantage ratio across both Qwen2.5-Math-7B and Qwen2.5-Math-1.5B. This reduction implies improved sample efficiency: a higher fraction of rollouts contributes meaningful learning signals.

\begin{figure}[!hbt]
    \centering
    \begin{subfigure}[t]{0.22\textwidth}
        \centering
        \includegraphics[width=\linewidth]{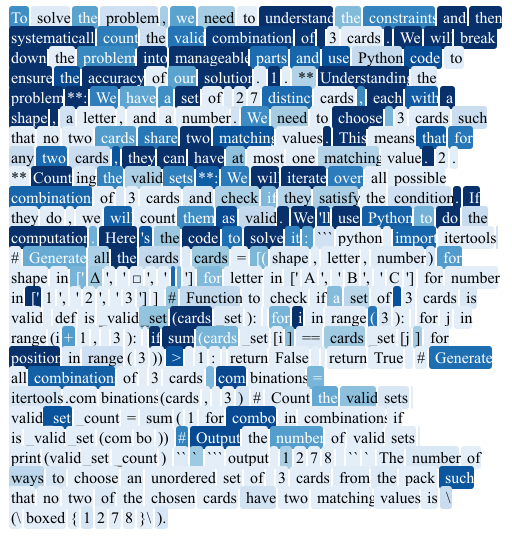}
        \caption{Correct answer \\\textcolor{red}{$\times$ 1.65}}
    \end{subfigure}
    ~
    \begin{subfigure}[t]{0.22\textwidth}
        \centering
        \includegraphics[width=\linewidth]{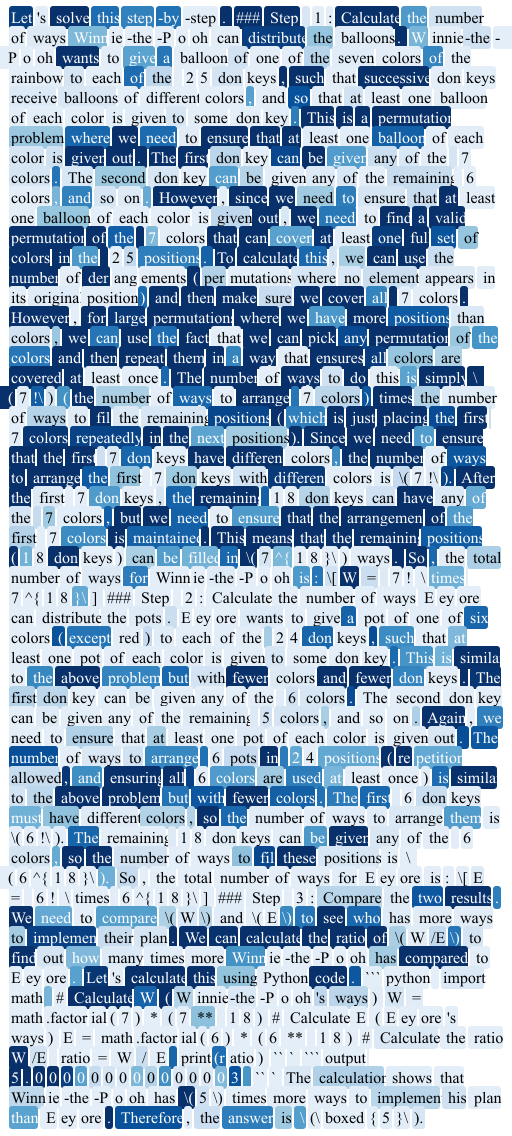}
        \caption{Incorrect answer \\\textcolor{red}{$\times$ 1.33}}
    \end{subfigure}
    ~ 
    \vspace{0.1cm}
    \begin{subfigure}[t]{0.22\textwidth}
        \centering
        \includegraphics[width=\linewidth]{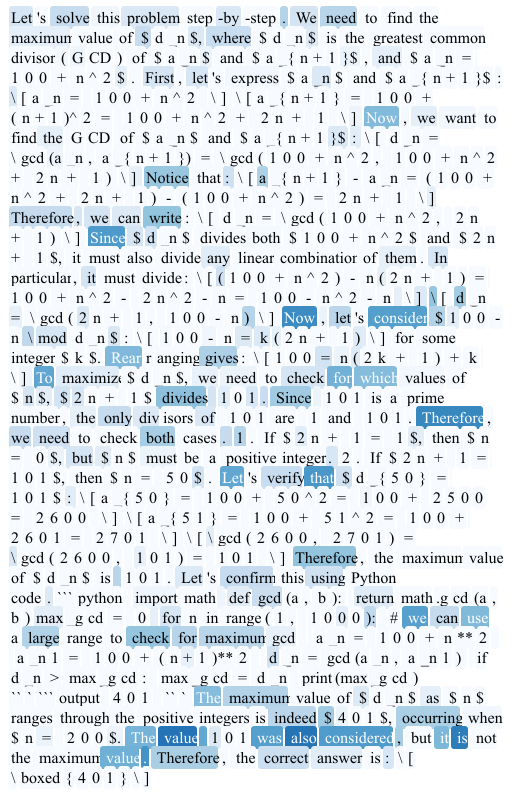}
        \caption{Correct answer \\\textcolor{red}{$\times$ 0.73}}
    \end{subfigure}
    ~ 
    \begin{subfigure}[t]{0.22\textwidth}
        \centering
        \includegraphics[width=\linewidth]{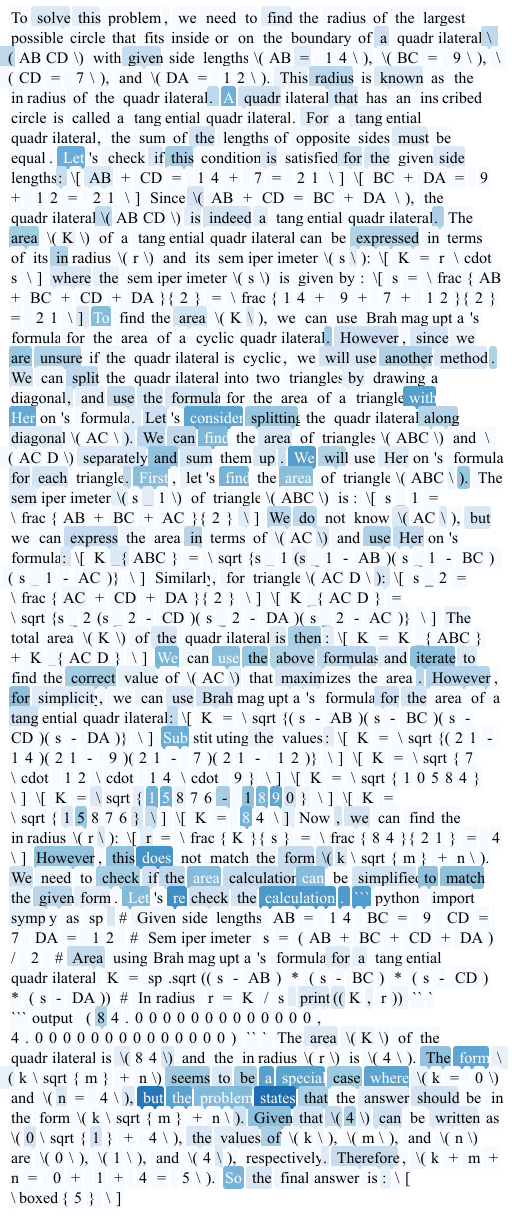}
        \caption{Incorrect answer \\\textcolor{red}{$\times$ 0.74}}
    \end{subfigure}
    \caption{Token-level entropy of sequences within a mini-batch (with relative-entropy in \textcolor{red}{red}).}
    \label{fig:rollout_entropy}
\end{figure}

\paragraph{Case Analysis.} To visualize the effect of relative entropy weighting, Figure~\ref{fig:rollout_entropy} displays examples of token-level entropy from rollouts within the same mini-batch, where darker shades indicate higher token entropy and the red multiplier denotes the relative entropy weight $w_i$. Rollouts with higher relative entropy (e.g., $\times$ 1.65) correspond to more exploratory reasoning paths, whereas lower-weight rollouts ($\times$ 0.73) reflect more confident, deterministic responses. This analysis highlights how PREPO adaptively balances learning signals that uncertain yet potentially informative rollouts receive stronger gradient influence, while confident but low-diversity outputs are proportionally downweighted.

\begin{figure}[!htb]
    \centering
    \begin{subfigure}[t]{0.45\textwidth}
        \centering
        \includegraphics[width=\linewidth]{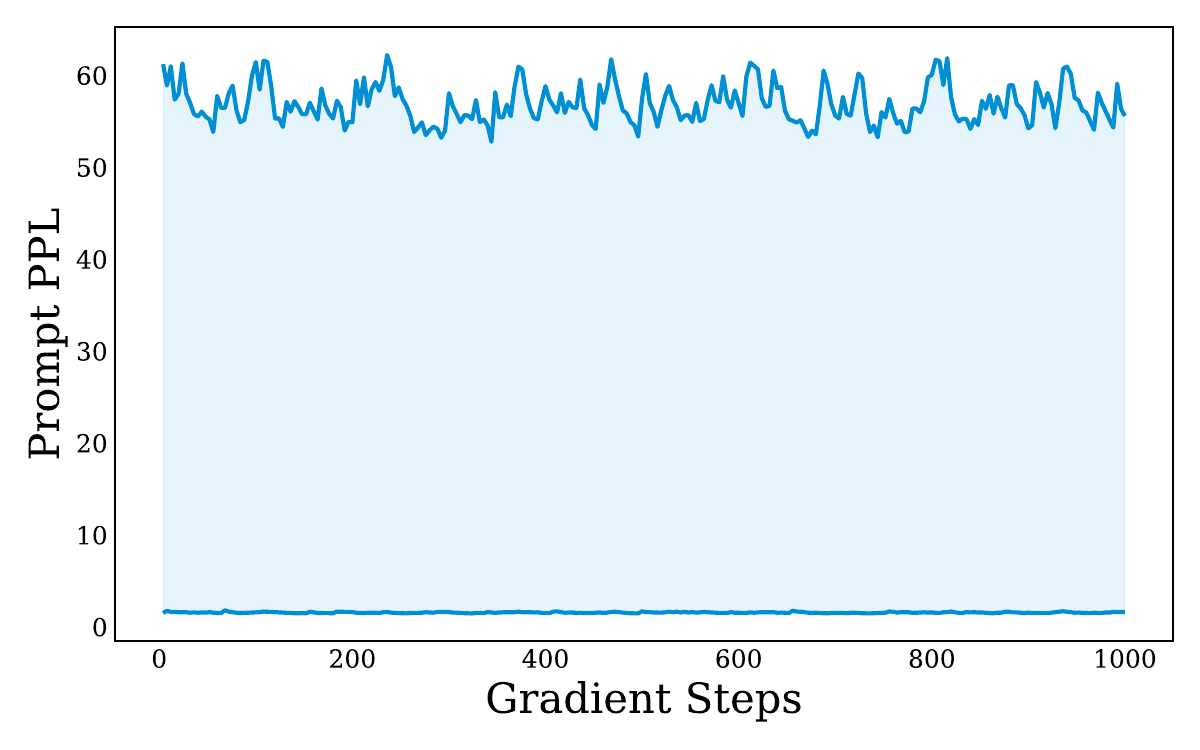}
        \caption{Qwen2.5-Math-7B}
    \end{subfigure}%
    ~ 
    \begin{subfigure}[t]{0.45\textwidth}
        \centering
        \includegraphics[width=\linewidth]{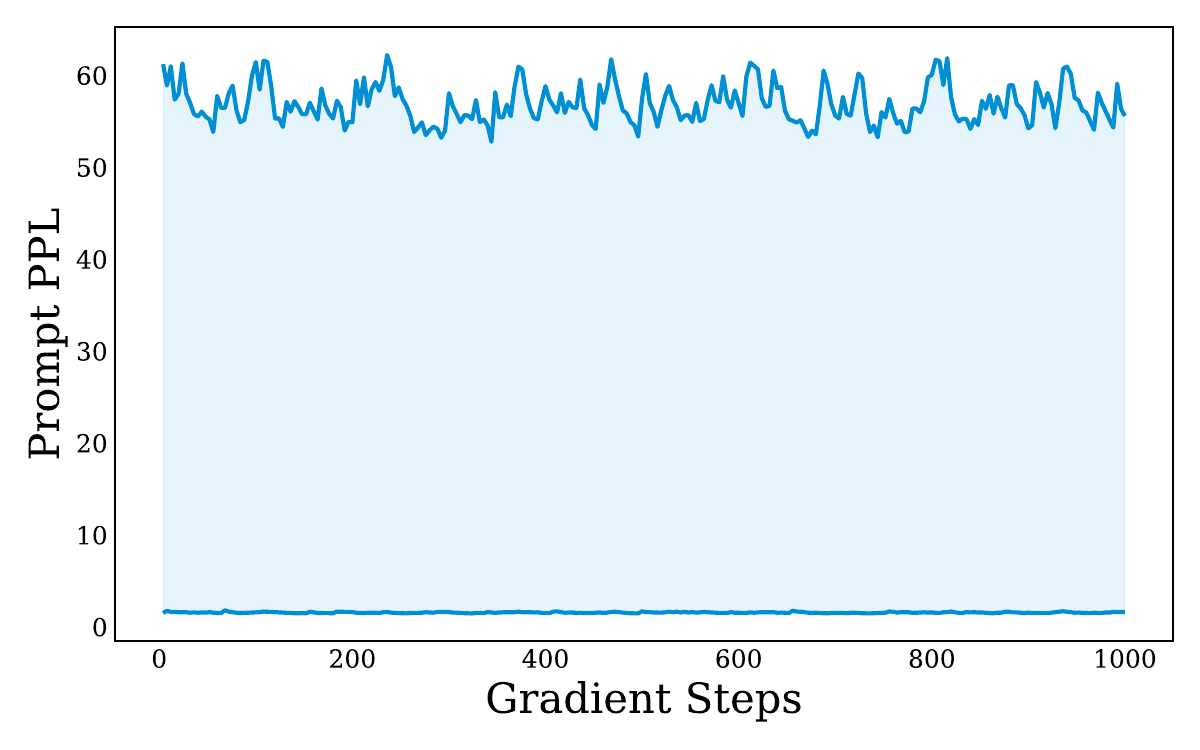}
        \caption{Qwen2.5-Math-1.5B} 
    \end{subfigure}
    \caption{Range of prompt PPL during training.}
    \label{fig:all_ppl}
\end{figure}

\paragraph{Does the prompt PPL vary during training?}
To examine whether the distribution of prompt difficulty shifts throughout training, we track the range of prompt PPL values at each epoch (Figure~\ref{fig:all_ppl}). The results show that the overall PPL range remains stable, and the mean prompt PPL exhibits minimal drift. This indicates that PREPO’s sampling strategy maintains a consistent distribution of task difficulty, preventing a bias toward easier prompts.

\begin{figure}[!htb]
    \centering
    \includegraphics[width=0.5\linewidth]{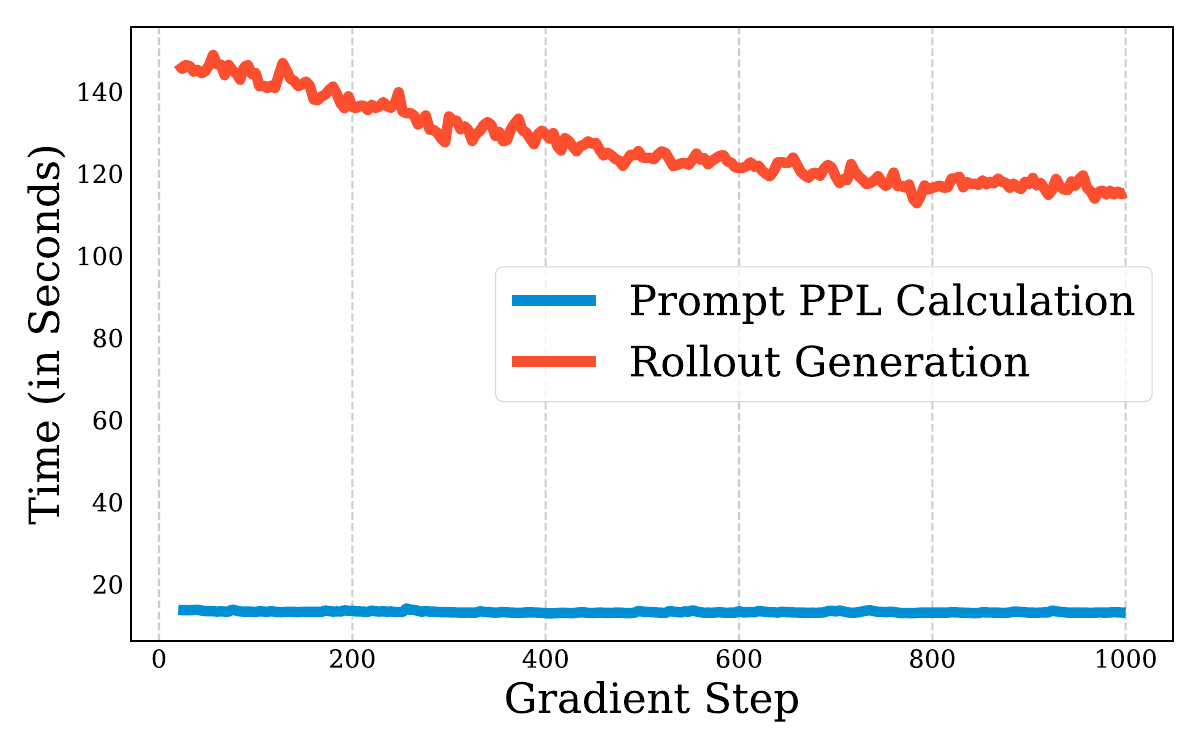}
    \caption{Comparison of calculating prompt PPL and rollout generation}
    \label{fig:cal_ppl_gen}
\end{figure}

\paragraph{Does PREPO affect training time per step?}
As illustrated in Figure~\ref{fig:cal_ppl_gen}, computing prompt PPL introduces only negligible overhead relative to rollout generation time. This confirms that PREPO’s filtering and weighting procedures are computationally efficient and thus scalable to larger models and datasets.

\begin{figure}[!bht]
    \centering
    \begin{subfigure}[t]{0.5\textwidth}
        \centering
       \includegraphics[width=\linewidth]{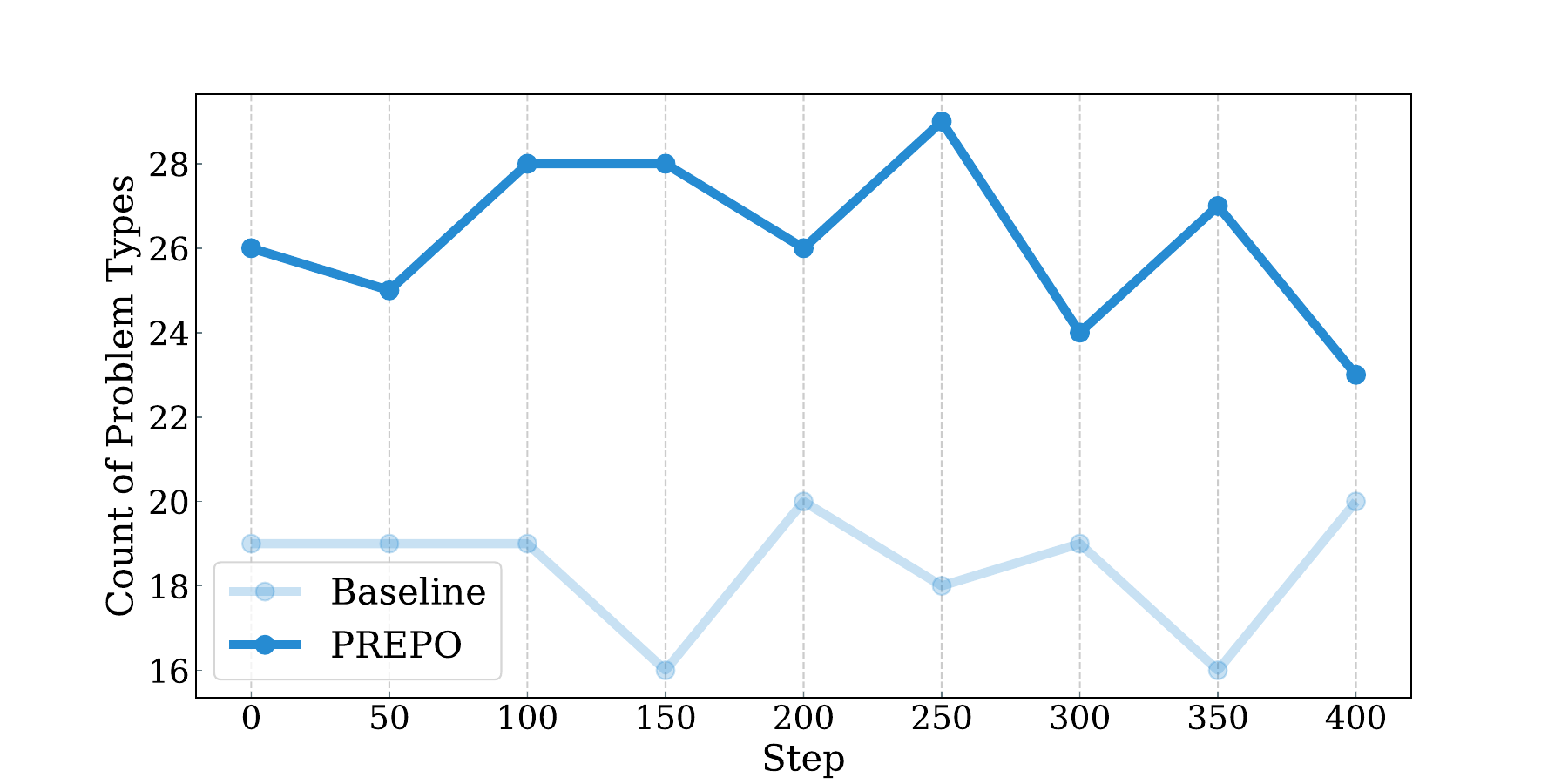}
        \caption{Unique number of MSC tags}
    \end{subfigure}
    ~~
    \vspace{0.1cm}
    \begin{subfigure}[t]{0.48\textwidth}
        \centering
        \includegraphics[width=\linewidth]{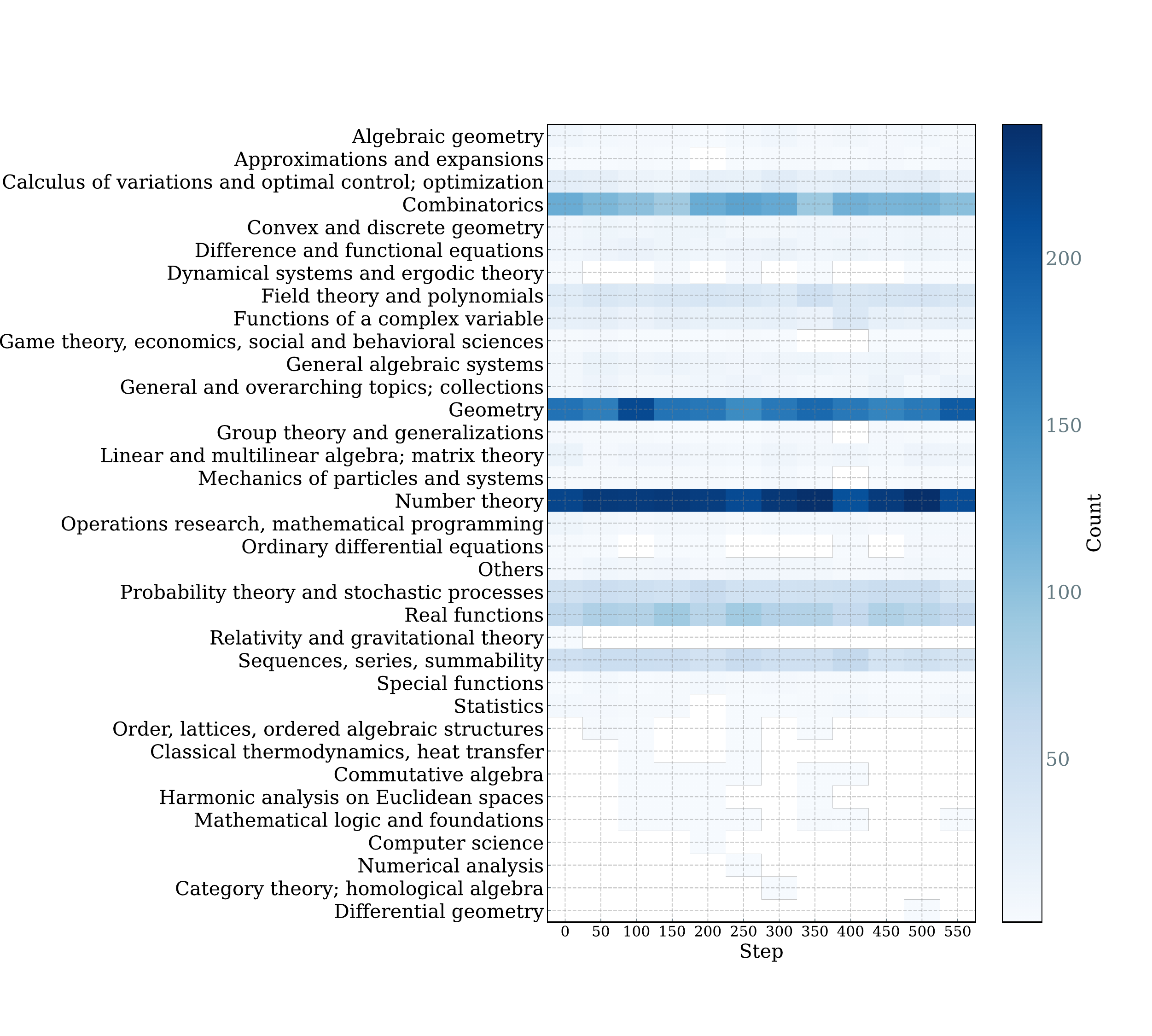}
        \caption{MSC frequency of PREPO}
    \end{subfigure}
    ~~
    \begin{subfigure}[t]{0.48\textwidth}
        \centering
        \includegraphics[width=\linewidth]{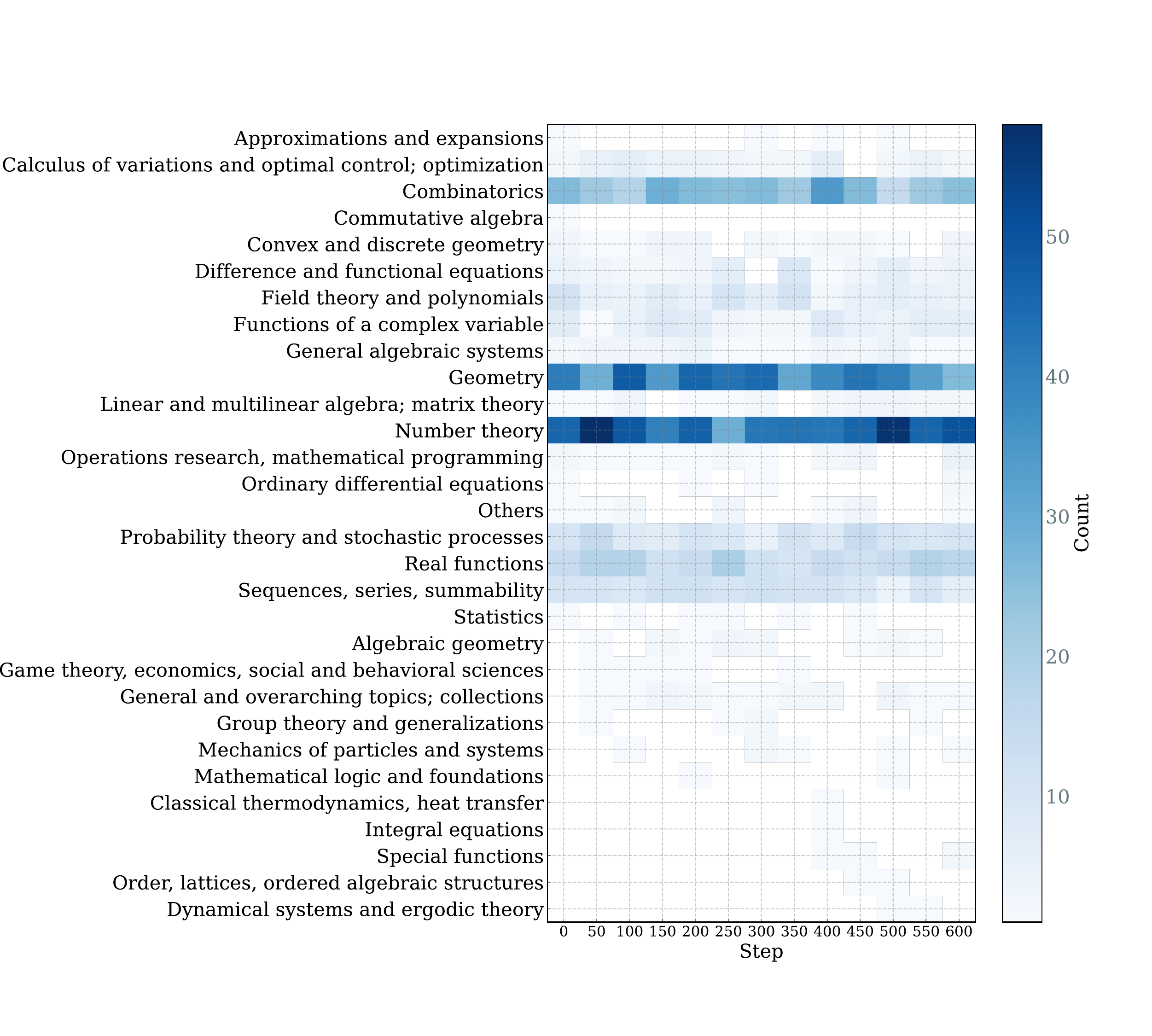}
        \caption{MSC frequency of baseline (random)}
    \end{subfigure}
    \caption{Comparison of problem diversity between PREPO and baseline on Qwen2.5-Math-7B.}
    \label{fig:problem_diveristy}
\end{figure}

\paragraph{Does PREPO select diverse problems?}
A further analysis of the training corpus reveals that PREPO samples a more diverse set of mathematical problems than random selection. As shown in Figure~\ref{fig:problem_diveristy}, PREPO achieves broader coverage across Mathematics Subject Classification (MSC\footnote{\href{https://zbmath.org/classification/}{https://zbmath.org/classification/}}) categories, indicating that its adaptive filtering promotes exposure to a wider range of reasoning types.

\begin{figure}[!bht]
    \centering
    \begin{subfigure}[t]{0.45\textwidth}
        \centering
        \includegraphics[width=\linewidth]{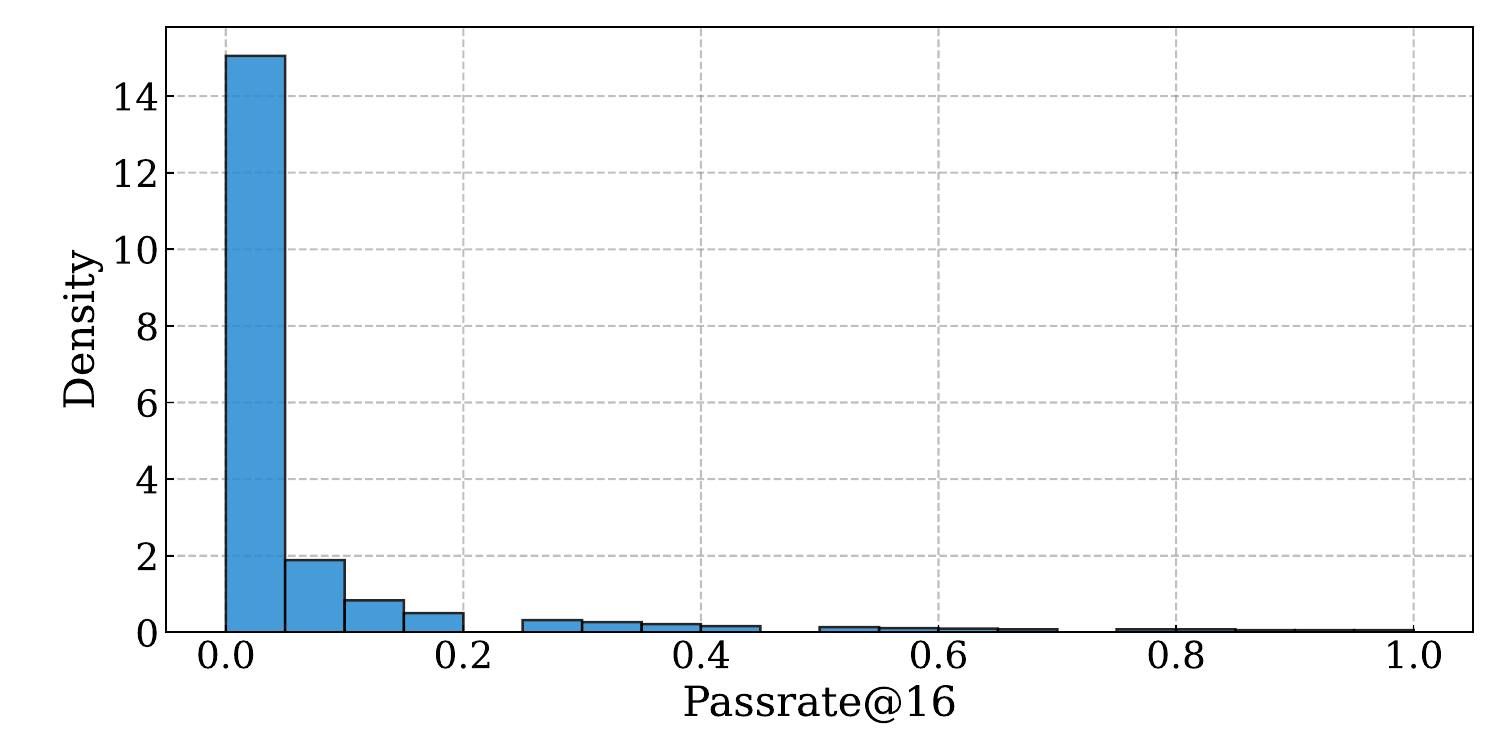}
        \caption{Qwen2.5-Math-1.5B}
    \end{subfigure}
    ~ 
    \begin{subfigure}[t]{0.45\textwidth}
        \centering
        \includegraphics[width=\linewidth]{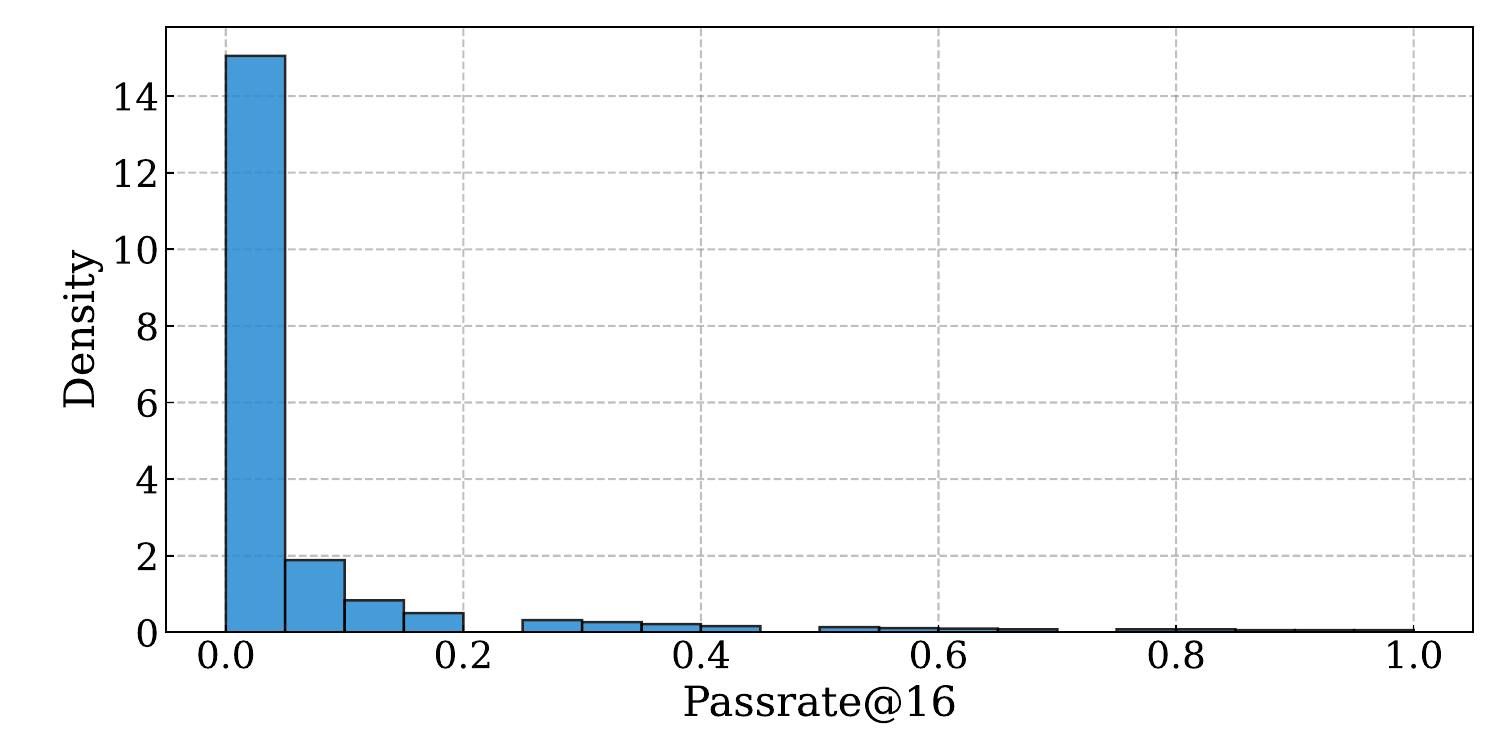}
        \caption{Qwen2.5-Math-7B}
    \end{subfigure}
    ~ 
    \begin{subfigure}[t]{0.45\textwidth}
        \centering
        \includegraphics[width=\linewidth]{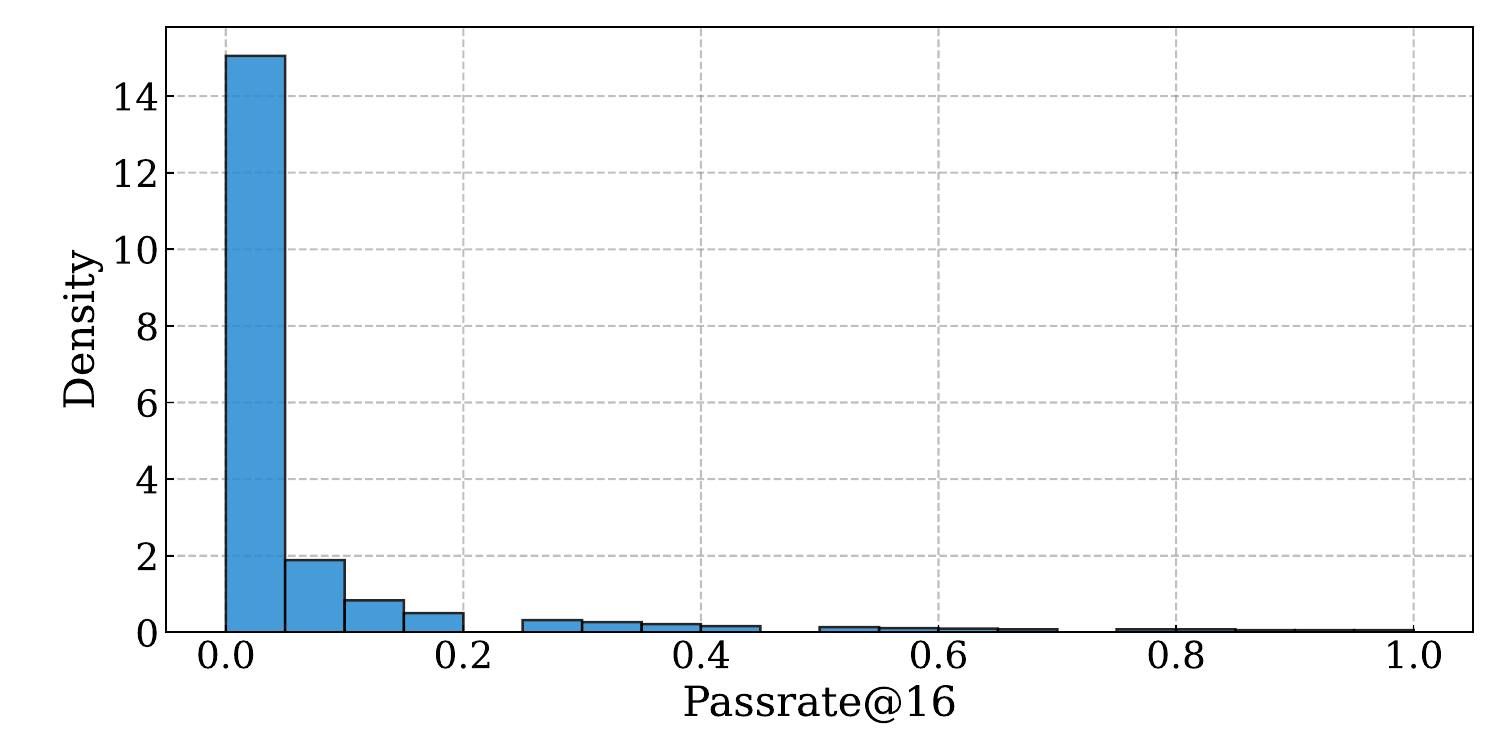}
        \caption{Qwen3-4B}
    \end{subfigure}
    ~
    \begin{subfigure}[t]{0.45\textwidth}
        \centering
        \includegraphics[width=\linewidth]{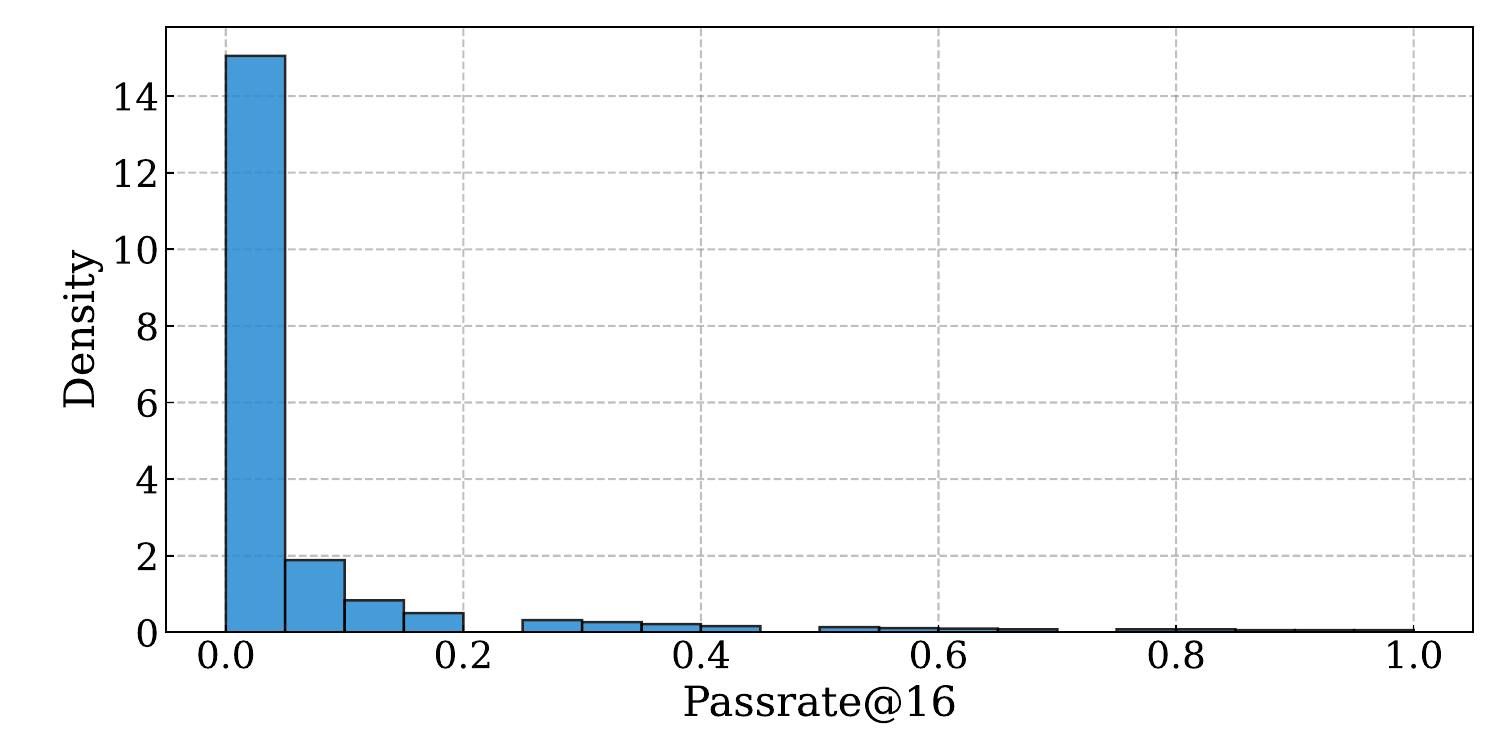}
        \caption{Llama3.1-8B}
    \end{subfigure}
    \caption{Distribution of passrate@16 over all partial prompts}
    \label{fig:partial_passrate}
\end{figure}

\paragraph{Does the model memorize the training data?}
To evaluate whether PREPO encourages memorization, we follow the methodology of \citet{wu2025reasoning} by truncating 40\% of each prompt to create partial problems. Models are then evaluated on these partial inputs, with performance measured by the average pass rate over 16 generations. As shown in Figure~\ref{fig:partial_passrate}, most partial problems yield near-zero pass rates, with only a small fraction achieving moderate success. This distribution implies that the models rely on full contextual information to solve tasks rather than recalling memorized solutions. Hence, PREPO’s improvements can be attributed to enhanced reasoning and generalization rather than rote memorization of training data.

\section{Conclusion and Future Work}
This study examined how intrinsic data properties can improve the efficiency of RLVR training. On the prompt side, we showed that prompt perplexity provides an effective indicator of model adaptability and naturally supports a data selection scheduling. On the rollout side, sequence-level entropy offered an intrinsic measure of response confidence. Building on these properties to guide exploration during training, we introduced PREPO, which integrates a perplexity-based schedule with entropy-based weighting. Empirical results from comprehensive experiments shows the effectiveness of PREPO. Future work may extend this direction by exploring additional intrinsic signals and integrating data-driven exploration with system-level optimization methods.

\section{Limitations}\label{app:limitation}
This study has several limitations that should be acknowledged.  
(1) The online selection ratio was fixed at 20\%, and the impact of alternative ratios has not been systematically examined;  
(2) response lengths were restricted to 32K tokens, leaving the applicability of PREPO to models generating substantially longer outputs an open question; and  
(3) the evaluation was limited to mathematical reasoning tasks, while its effectiveness in other domains remains to be explored.


\newpage
\bibliography{references}

\newpage
\appendix


\newpage
\appendix

\section{Entropy as a Confidence Signal in Rollouts}\label{app:entropy_motivation}

While prompt filtering determines the distribution of prompts ${x_i}$ presented to the model, standard RLVR methods rely solely on reward feedback and lack an explicit mechanism for assessing the confidence of generated responses. Entropy serves as a common measure of confidence, as it quantifies the sharpness of the predictive distribution along a rollout.

The token-level entropy of a rollout is defined as $H_t = - \sum_{v \in \mathcal{V}} \pi_\theta(v \mid o_{<t}, x)\log \pi_\theta(v \mid o_{<t}, x),$ where $\mathcal{V}$ is the vocabulary. The sequence-level entropy is then the average
\begin{equation}
\bar H(o \mid x) = \frac{1}{|o|} \sum_{t=1}^{|o|} H_t.
\end{equation}
Rollouts with low entropy correspond to highly concentrated predictive distributions, reflecting strong model confidence in a single continuation. 
In contrast, rollouts with high entropy correspond to more diffuse distributions, where multiple continuations remain plausible. 

Thus, entropy complements prompt filtering by providing an intrinsic measure of response confidence that can be directly incorporated into training.

\section{Additional Results on Training Dynamics of Low- and High-PPL Prompts}\label{app:low_high_prompt_dynamics}

\begin{figure}[!h]
    \centering
    \includegraphics[width=\linewidth]{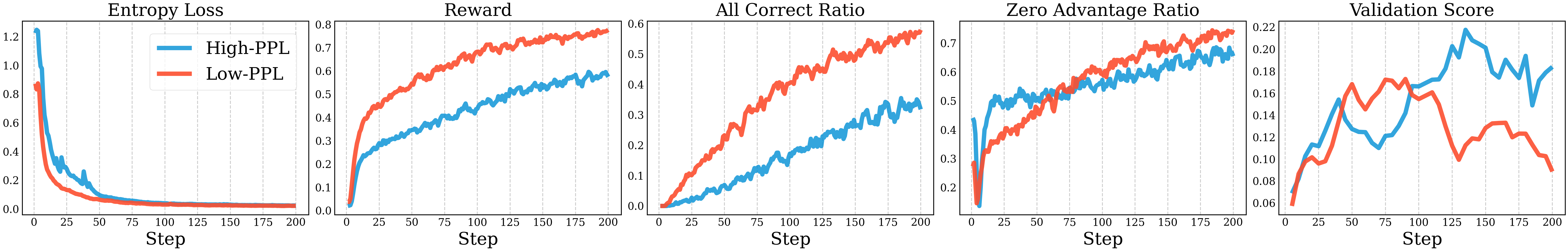}
    \caption{Training dynamics of \textsc{Low-PPL} vs. \textsc{High-PPL} prompts on Qwen2.5-7B.}
    \label{fig:pilot_study_7b}
\end{figure}

\begin{figure}[!h]
    \centering
    \includegraphics[width=\linewidth]{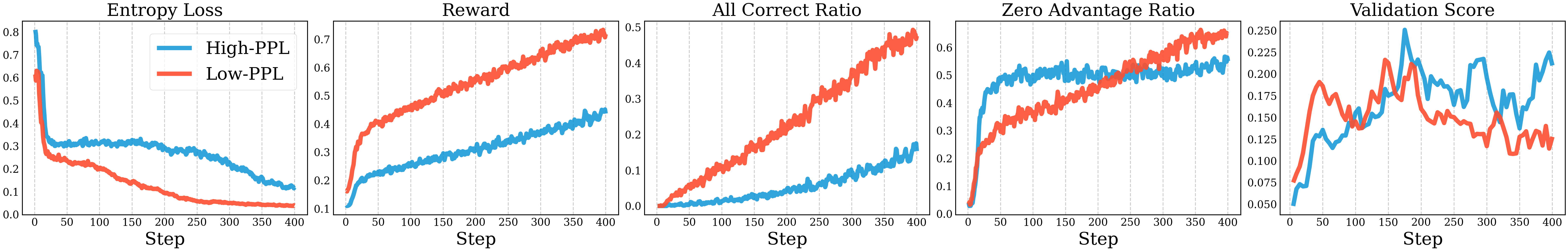}
    \caption{Training dynamics of \textsc{Low-PPL} vs. \textsc{High-PPL} prompts on Qwen2.5-Math-1.5B.}
    \label{fig:pilot_study_math_1.5b}
\end{figure}

Across Qwen models (Figure \ref{fig:pilot_study_7b} and Figure \ref{fig:pilot_study_math_1.5b}), we observe the following patterns:
\begin{itemize}[leftmargin=*]
\item \textbf{Entropy.} Prompts with high perplexity (High-PPL) have consistently higher entropy compared to those with low perplexity (Low-PPL).
\item \textbf{Reward.} Low-PPL prompts receive higher reward values compaired with the High-PPL group.
\item \textbf{All-Correct Ratio.} Low-PPL prompts reach saturation more rapidly, with a larger proportion of prompts producing fully correct responses.
\item \textbf{Validation Score (AIME24 \citep{aime2024I}).} The Low-PPL group also achieves superior validation performance in the early training stages, suggesting that exposure to more “familiar’’ data facilitates faster adaptation and knowledge acquisition in large language models.
\item \textbf{Zero-Advantage Ratio.} In later training phases, the Low-PPL group has a higher zero-variance ratio, resulting in fewer effective rollouts and reduced sample efficiency compared to the High-PPL group.
\end{itemize}

\begin{figure}[!h]
    \centering
    \includegraphics[width=\linewidth]{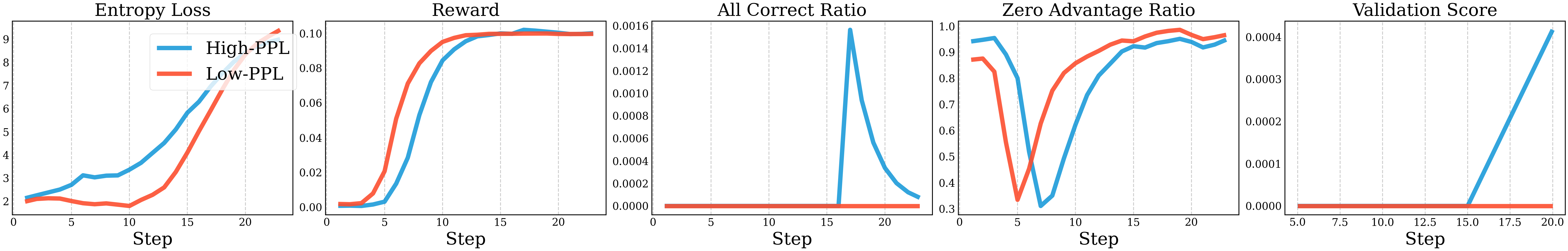}
    \caption{Training dynamics of \textsc{Low-PPL} vs. \textsc{High-PPL} prompts on Llama3.1-8B.}
    \label{fig:pilot_study_math_l8b}
\end{figure}

For Llama3.1-8B (Figure~\ref{fig:pilot_study_math_l8b}), we similarly observe that High-PPL prompts lead to higher entropy and lower reward curves. However, overall performance remains poor because the dataset is too challenging for Llama3.1-8B to be effectively trained with RLVR.

\section{Visualization of High-PPL and Low-PPL Prompts}

\begin{figure}[!htb]
    \centering
    \begin{subfigure}[t]{0.4\textwidth}
        \centering
        \includegraphics[width=\linewidth]{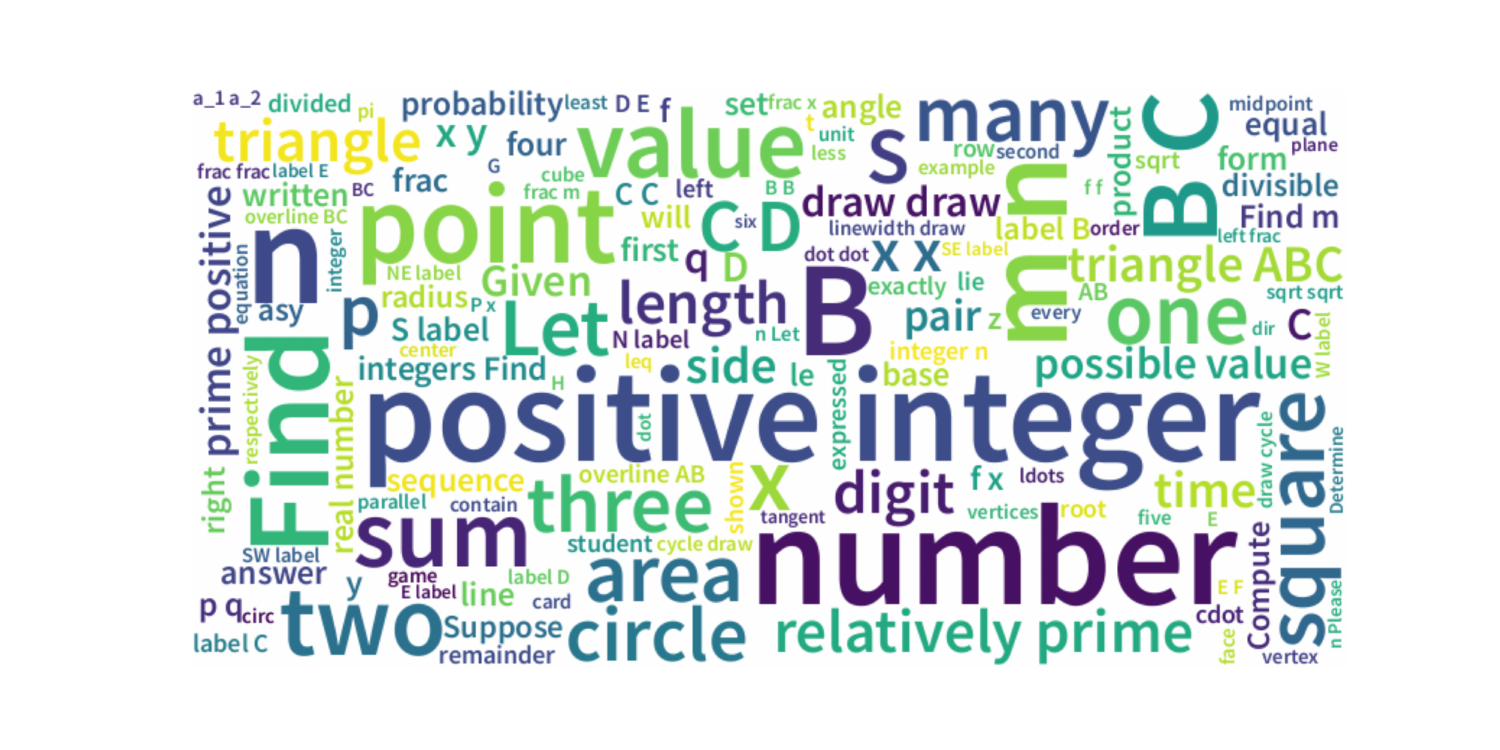}
        \caption{Low-PPL prompts (Qwen2.5-7B) }
    \end{subfigure}%
    ~ 
    \begin{subfigure}[t]{0.4\textwidth}
        \centering
        \includegraphics[width=\linewidth]{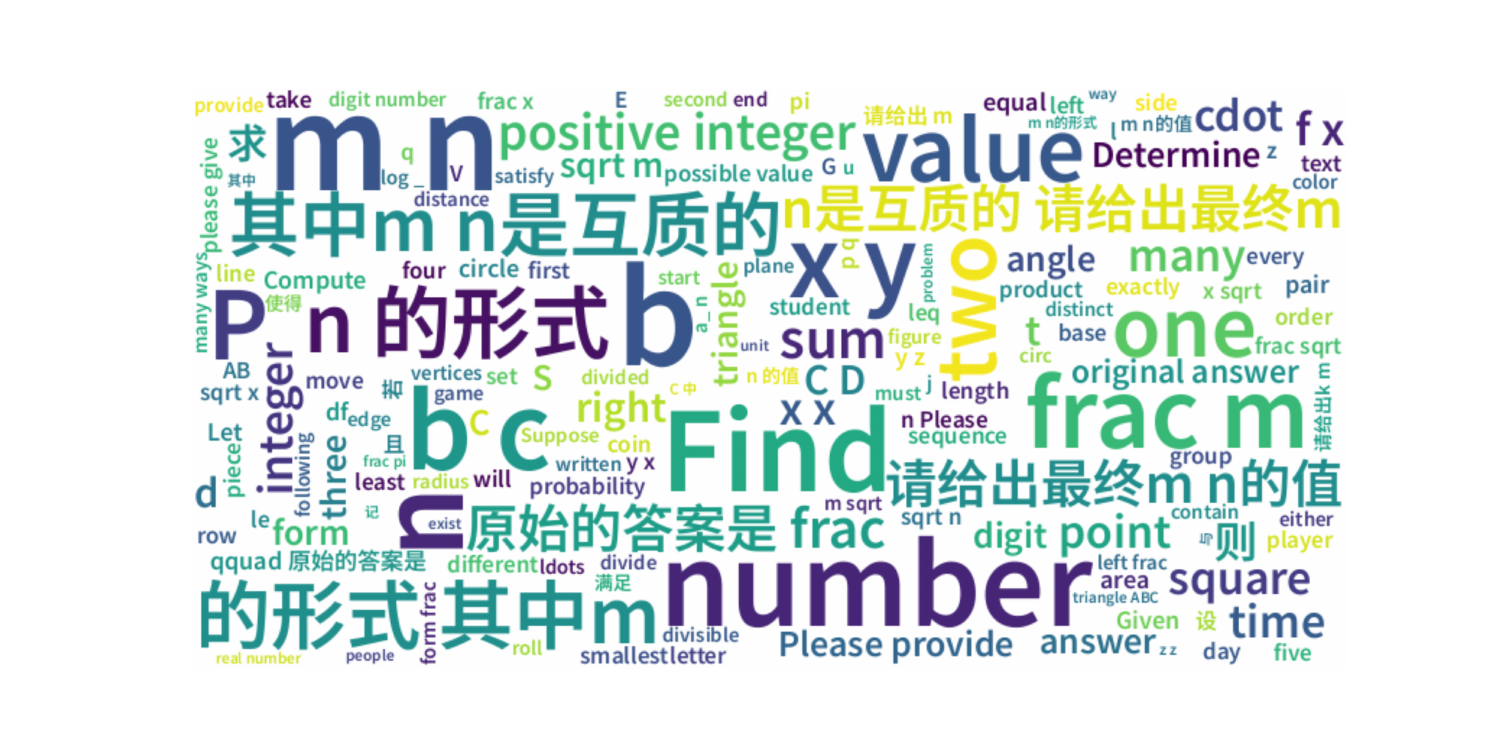}
        \caption{High-PPL prompts (Qwen2.5-7B)}
    \end{subfigure}
    ~
    \begin{subfigure}[t]{0.4\textwidth}
        \centering
        \includegraphics[width=\linewidth]{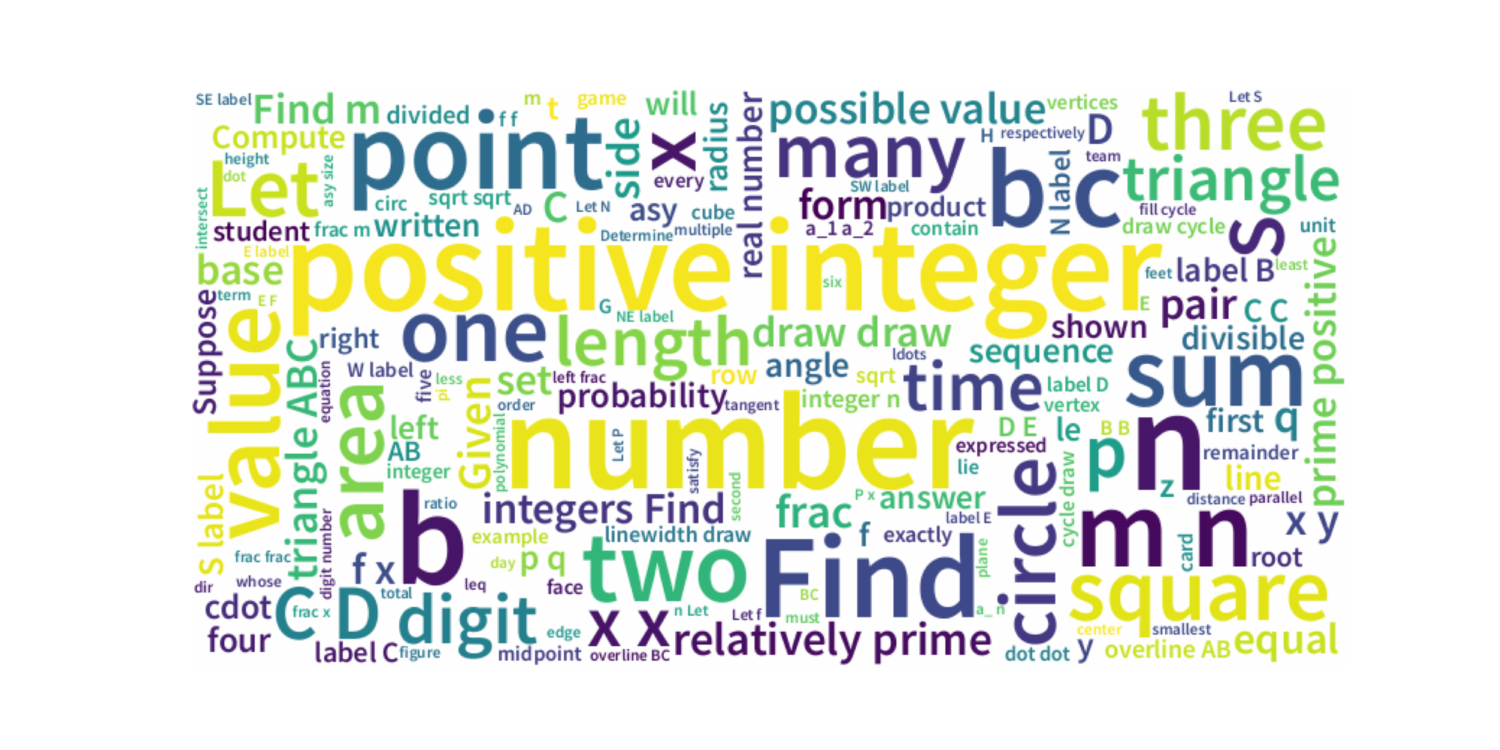}
        \caption{Low-PPL prompts (Qwen2.5-Math-1.5B) }
    \end{subfigure}%
    ~
    \begin{subfigure}[t]{0.4\textwidth}
        \centering
        \includegraphics[width=\linewidth]{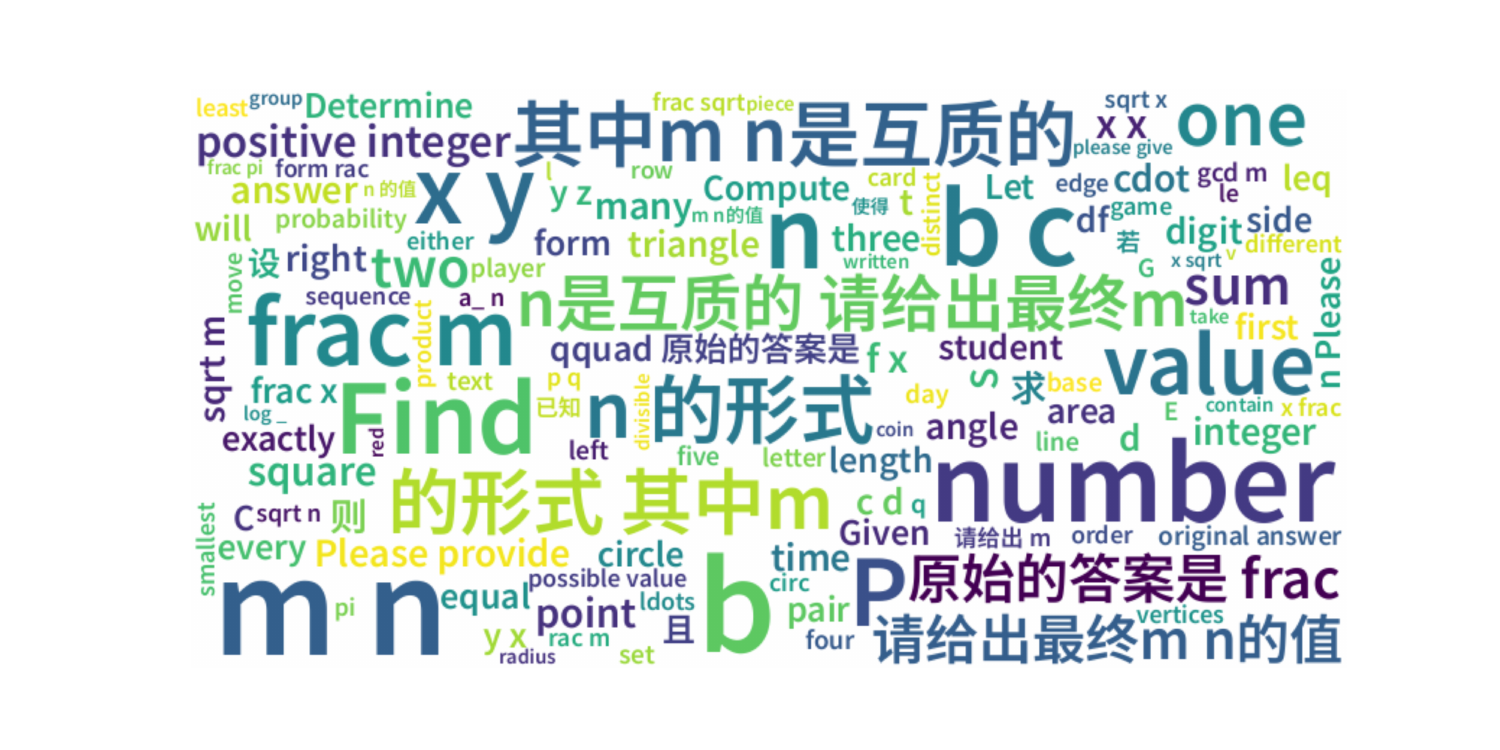}
        \caption{High-PPL prompts (Qwen2.5-Math-1.5B)}
    \end{subfigure}
    ~
    \begin{subfigure}[t]{0.4\textwidth}
        \centering
        \includegraphics[width=\linewidth]{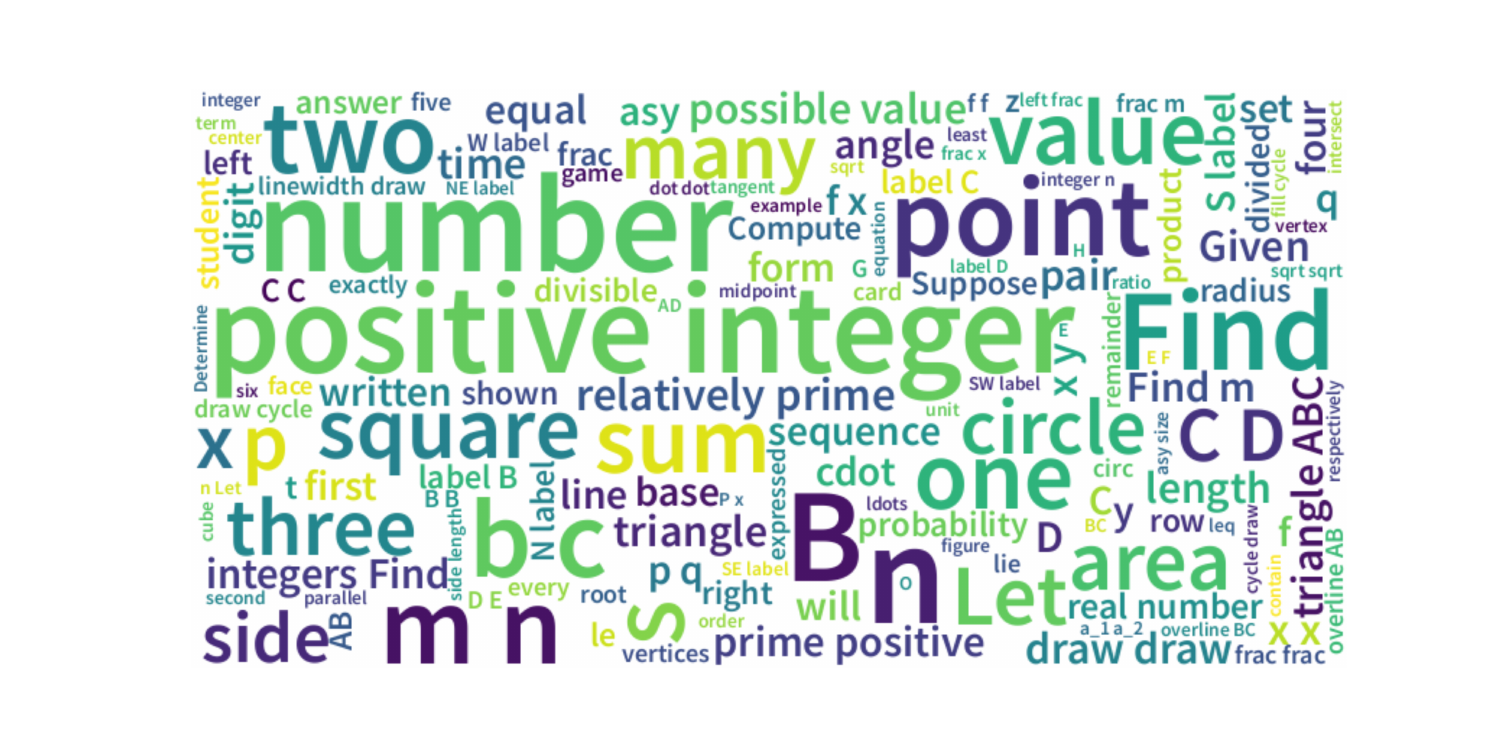}
        \caption{Low-PPL prompts (Qwen2.5-Math-7B) }
    \end{subfigure}%
    ~ 
    \begin{subfigure}[t]{0.4\textwidth}
        \centering
        \includegraphics[width=\linewidth]{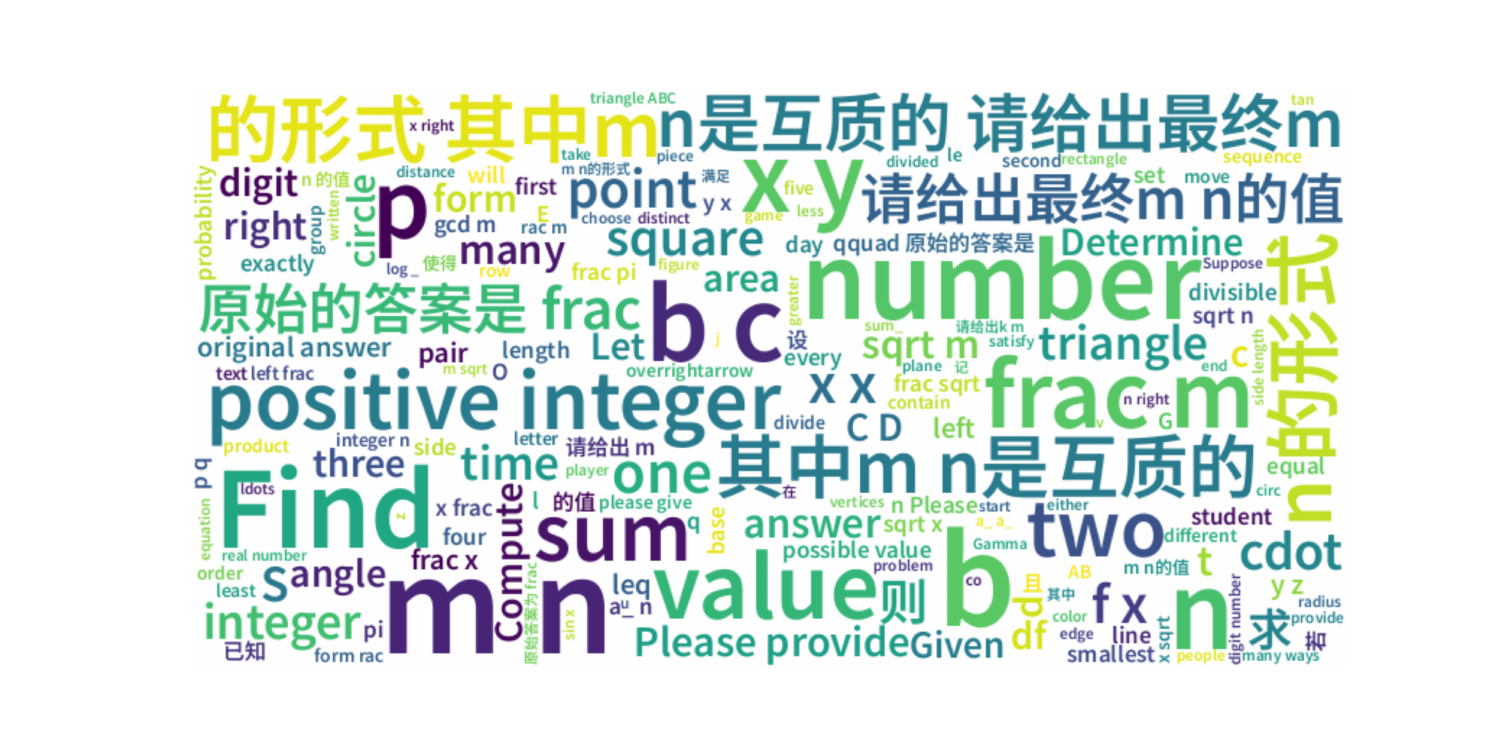}
        \caption{High-PPL prompts (Qwen2.5-Math-7B)}
    \end{subfigure}
    \caption{Wordcloud of the most frequent words in Low-/High-PPL prompts}
    ~
    \begin{subfigure}[t]{0.4\textwidth}
        \centering
        \includegraphics[width=\linewidth]{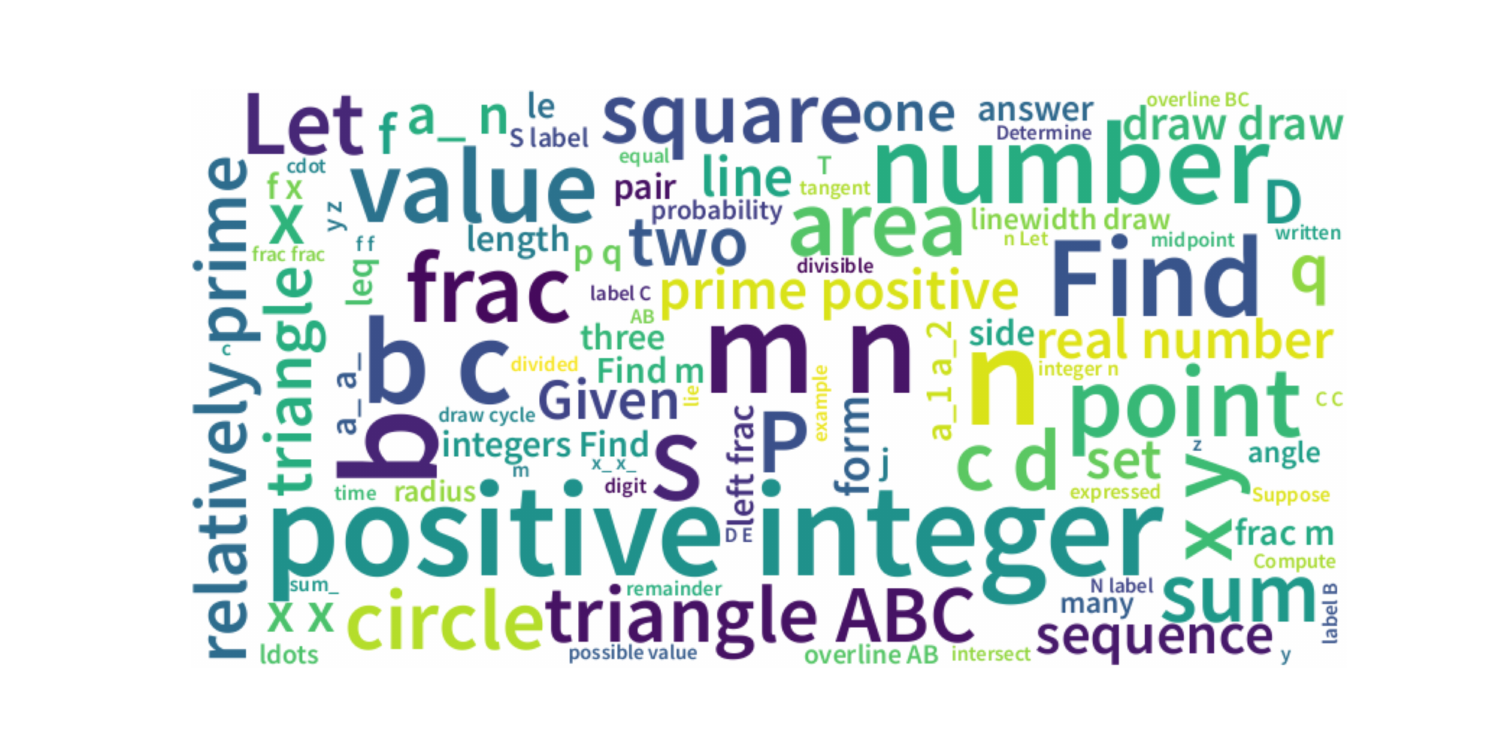}
        \caption{Low-PPL prompts (Llama3.1-8B) }
    \end{subfigure}%
    ~ 
    \begin{subfigure}[t]{0.4\textwidth}
        \centering
        \includegraphics[width=\linewidth]{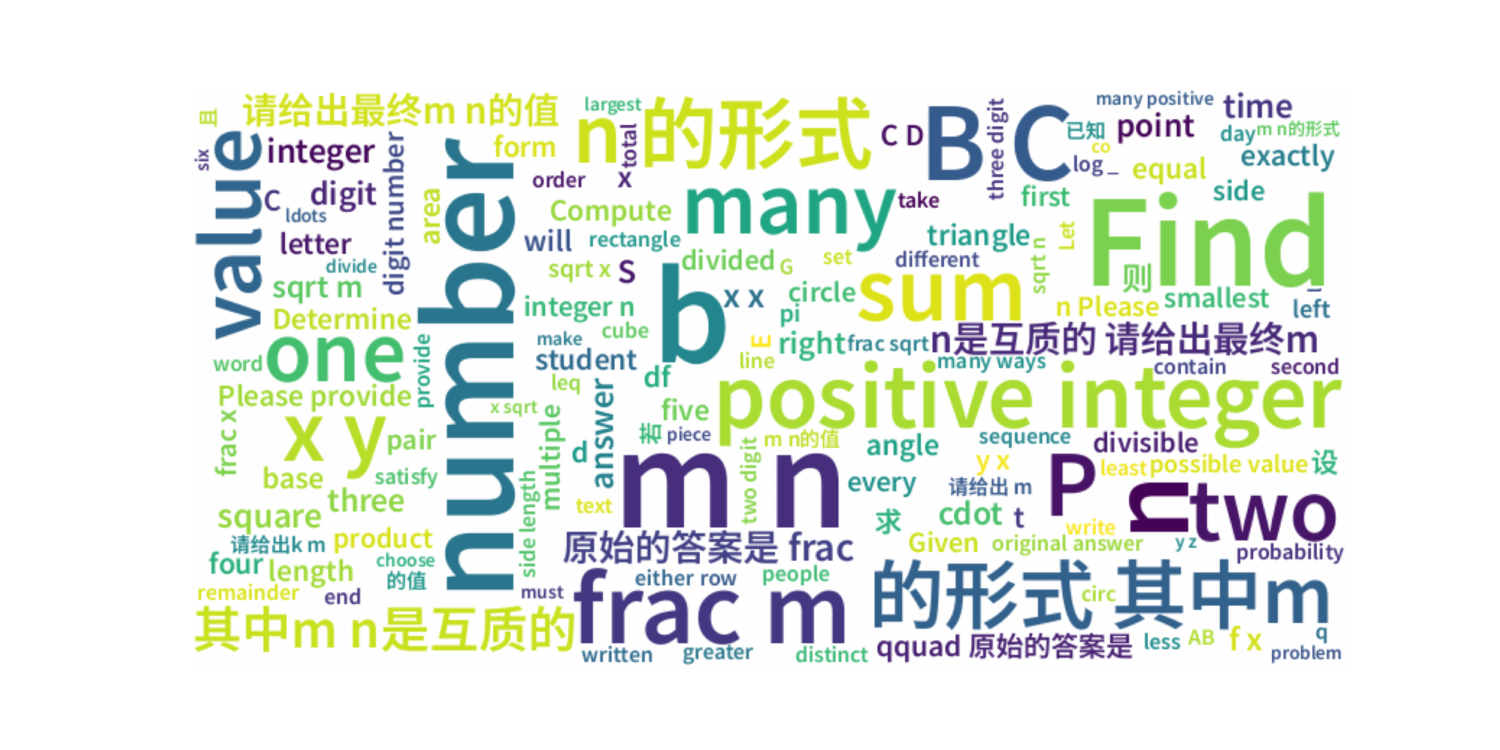}
        \caption{High-PPL prompts (Llama3.1-8B)}
    \end{subfigure}
    \caption{Wordcloud of the most frequent words in Low-/High-PPL prompts}
    \label{fig:wordcloud}
\end{figure}

As illustrated in Figure \ref{fig:wordcloud}, High-PPL prompts exhibit a greater prevalence of non-English characters relative to Low-PPL prompts. This pattern is consistently observed across both the Qwen2.5-series and Llama model families.

\section{Additional Results}\label{app:additional_experiments}
\subsection{Training Dynamics of PREPO versus the Baseline}

\begin{figure}[H]
    \centering
    \begin{subfigure}[t]{0.24\textwidth}
        \centering
        \includegraphics[width=\linewidth]{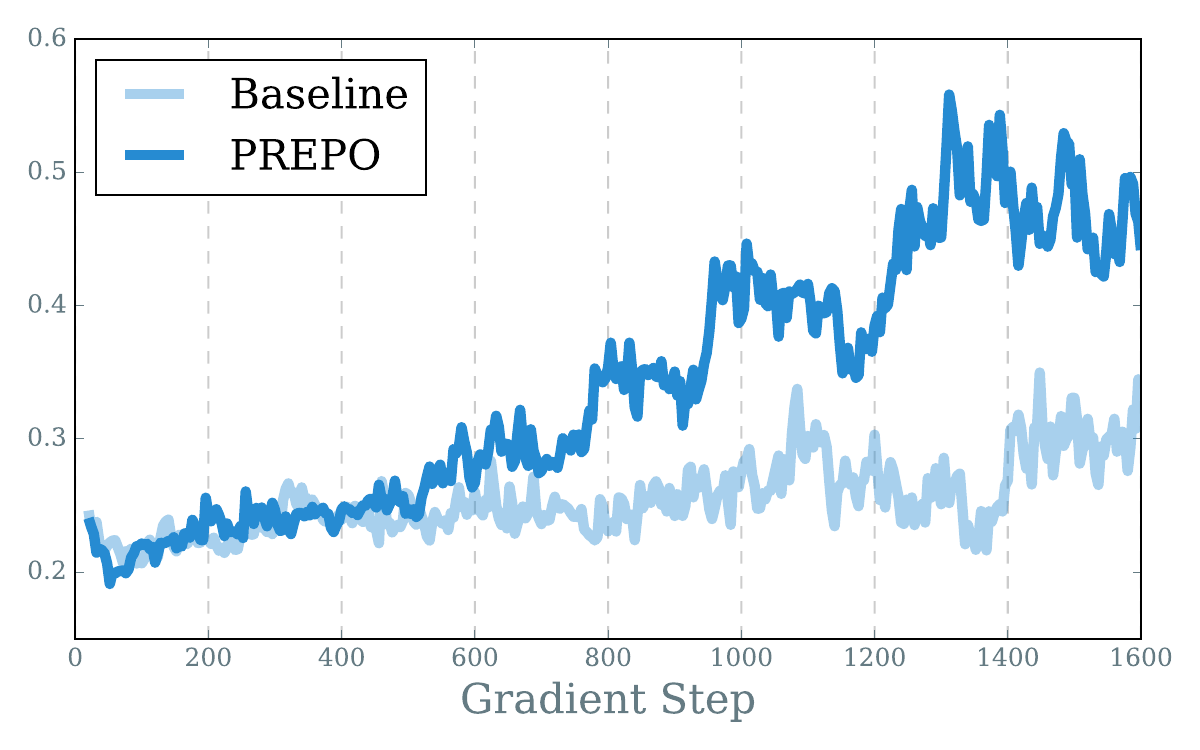}
        \caption{Entropy}
    \end{subfigure}
    \begin{subfigure}[t]{0.24\textwidth}
        \centering
        \includegraphics[width=\linewidth]{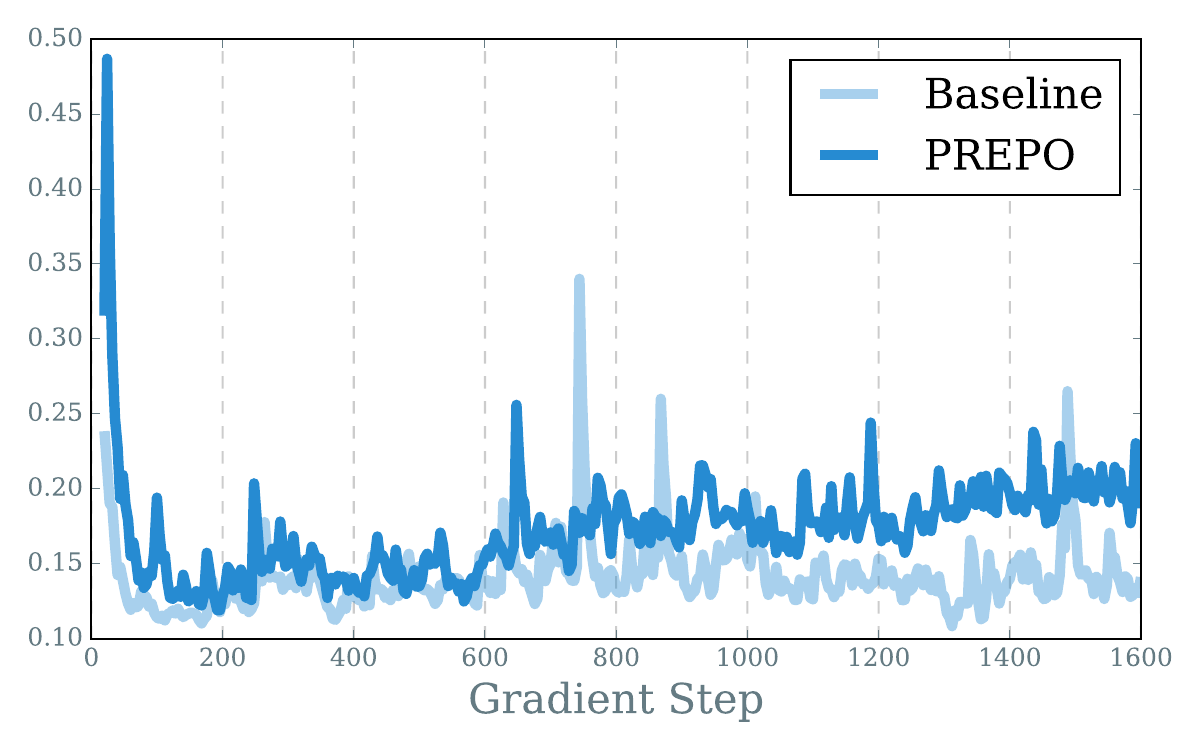}
        \caption{Gradient norm}
    \end{subfigure}
    \begin{subfigure}[t]{0.24\textwidth}
        \centering
        \includegraphics[width=\linewidth]{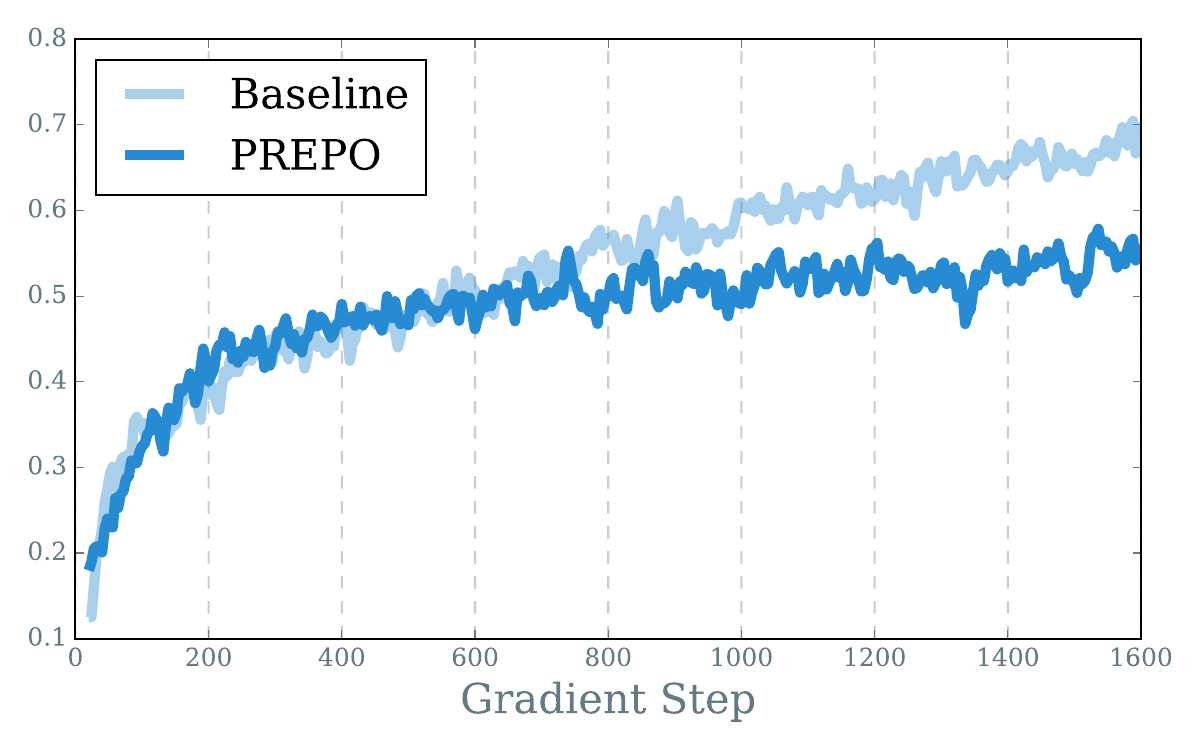}
        \caption{Zero adv. ratio}
    \end{subfigure}
    \begin{subfigure}[t]{0.24\textwidth}
        \centering
        \includegraphics[width=\linewidth]{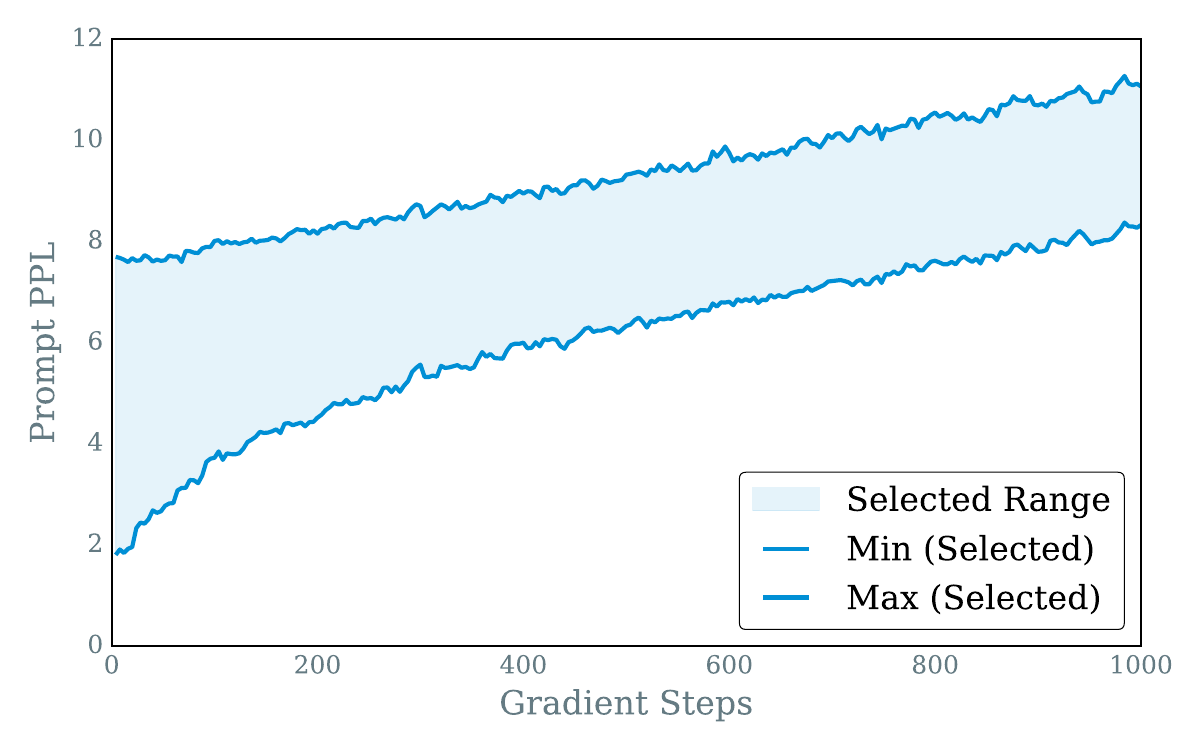}
        \caption{PREPO PPL range}
    \end{subfigure}
    \caption{Full Comparison between PREPO and random selection on Qwen2.5-Math-7B.}
    \label{fig:full_compare}
\end{figure}

\begin{figure}[H]
    \centering
    \begin{subfigure}[t]{0.24\textwidth}
        \centering
        \includegraphics[width=\linewidth]{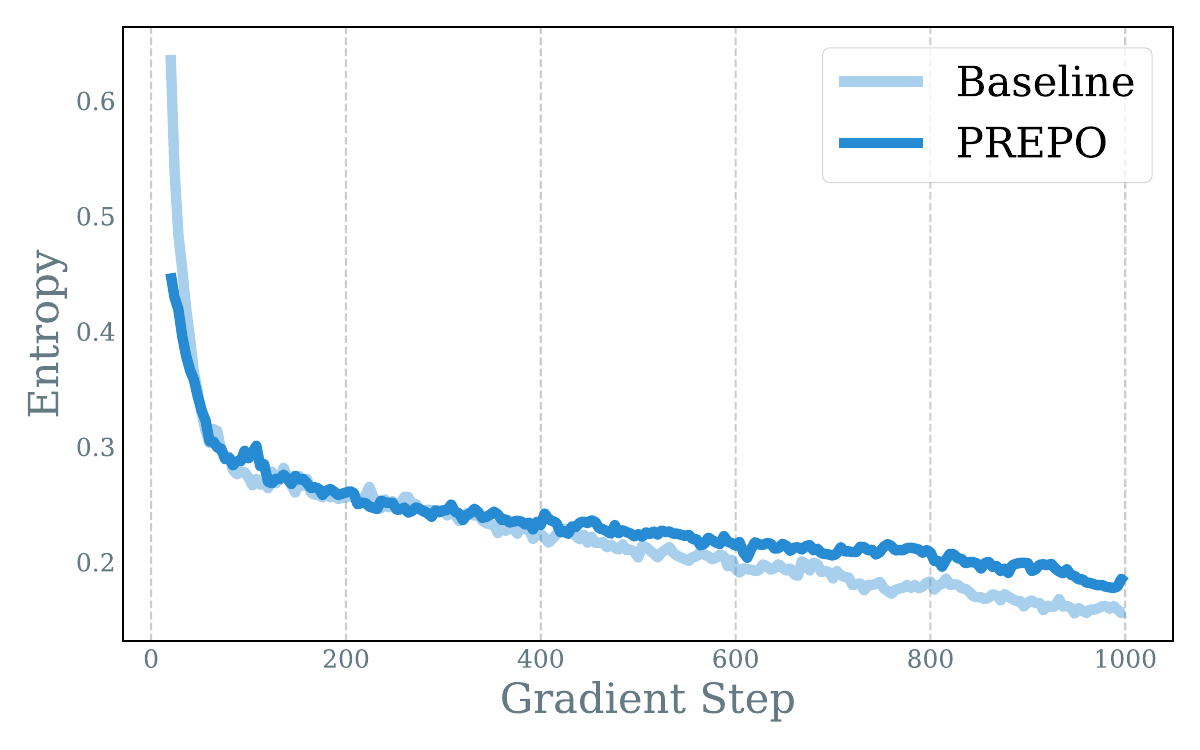}
        \caption{Entropy}
    \end{subfigure}
    \begin{subfigure}[t]{0.24\textwidth}
        \centering
        \includegraphics[width=\linewidth]{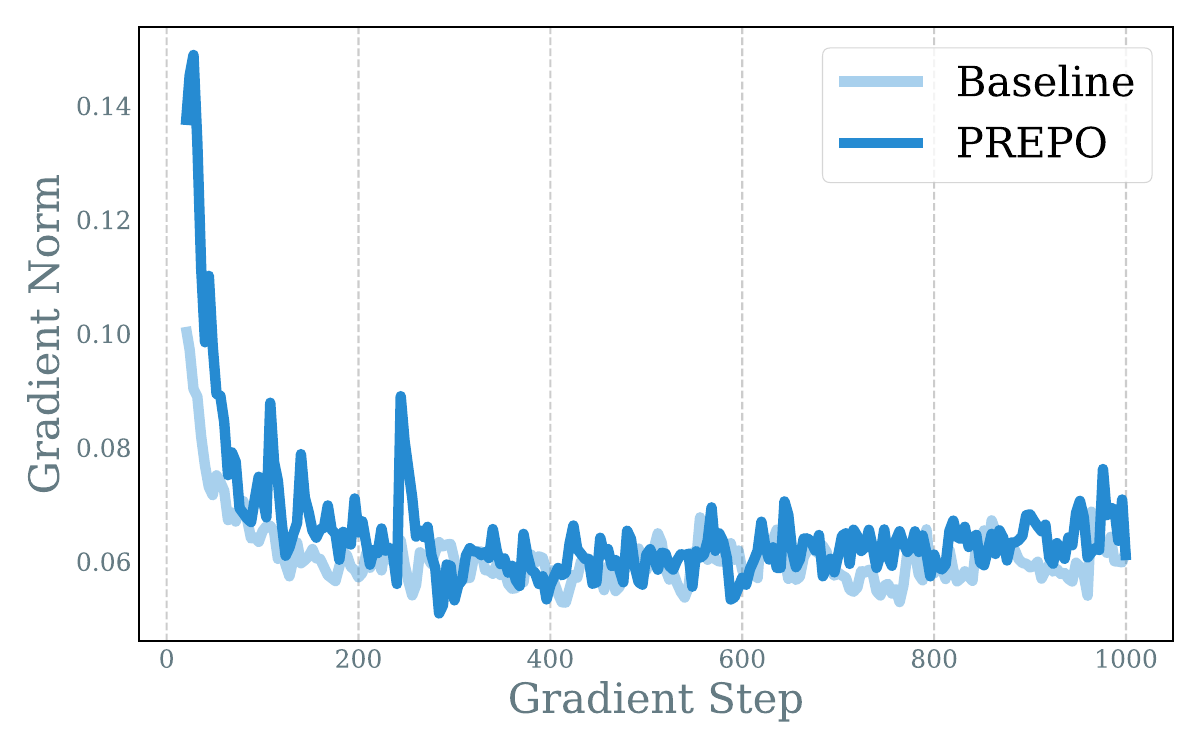}
        \caption{Gradient norm}
    \end{subfigure}
    \begin{subfigure}[t]{0.24\textwidth}
        \centering
        \includegraphics[width=\linewidth]{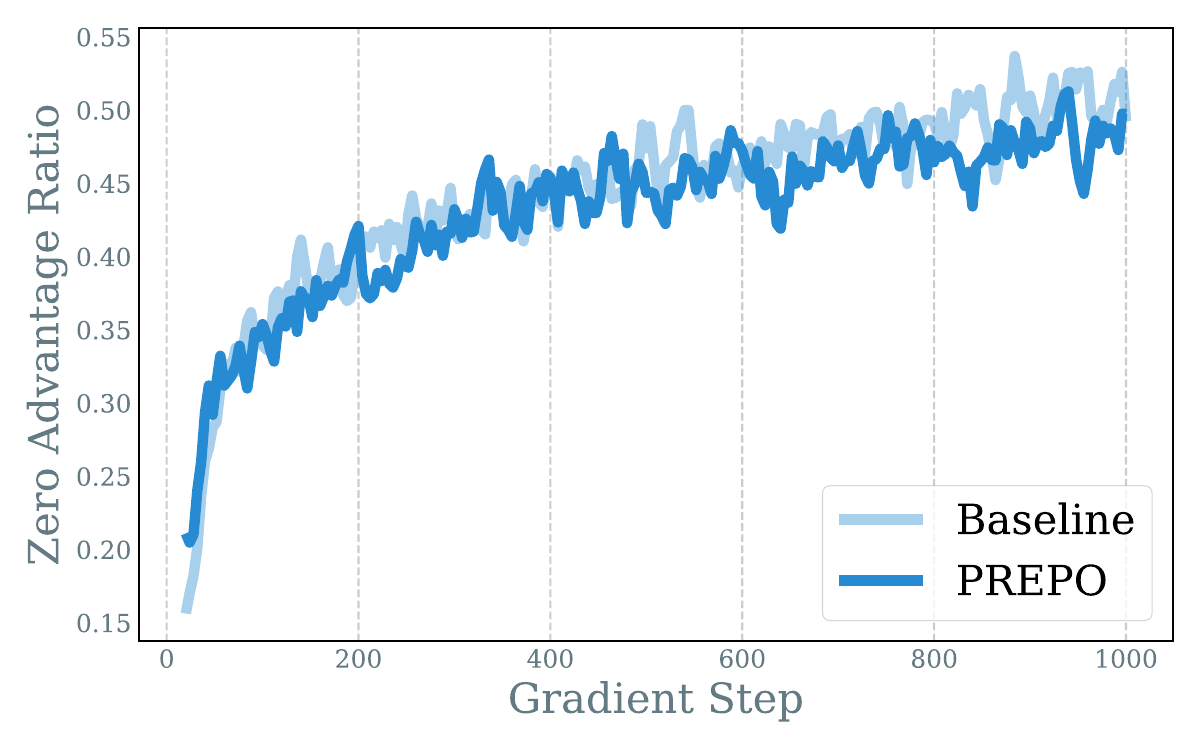}
        \caption{Zero adv. ratio}
    \end{subfigure}
    \begin{subfigure}[t]{0.24\textwidth}
        \centering
        \includegraphics[width=\linewidth]{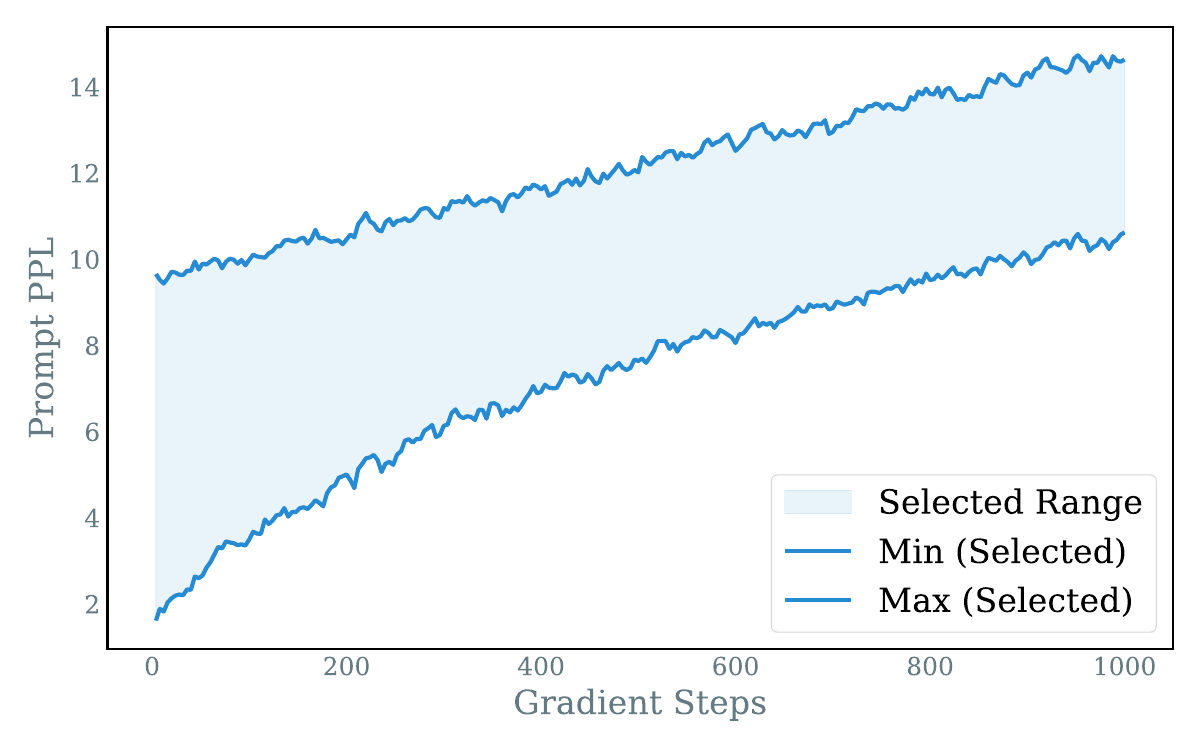}
        \caption{PREPO PPL range}
    \end{subfigure}
    \caption{Full Comparison between PREPO and random selection on Qwen2.5-Math-1.5B}
    \label{fig:full_compare_1_5b}
\end{figure}

As shown in Figure \ref{fig:full_compare} and \ref{fig:full_compare_1_5b}, both PREPO and the baseline show increasing entropy, though the rise is more pronounced under PREPO, indicating stronger exploration. PREPO also maintains a slightly higher gradient norm without instability. Moreover, it produces a smaller proportion of rollouts with zero advantage, suggesting that its rollouts yield more informative learning signals. Finally, the prompt perplexity selected by PREPO increases gradually throughout training, reflecting a consistent learning progression.

\subsection{Comparison with No-Filtering}
\begin{figure}[H]
    \centering
    \includegraphics[width=0.9\linewidth]{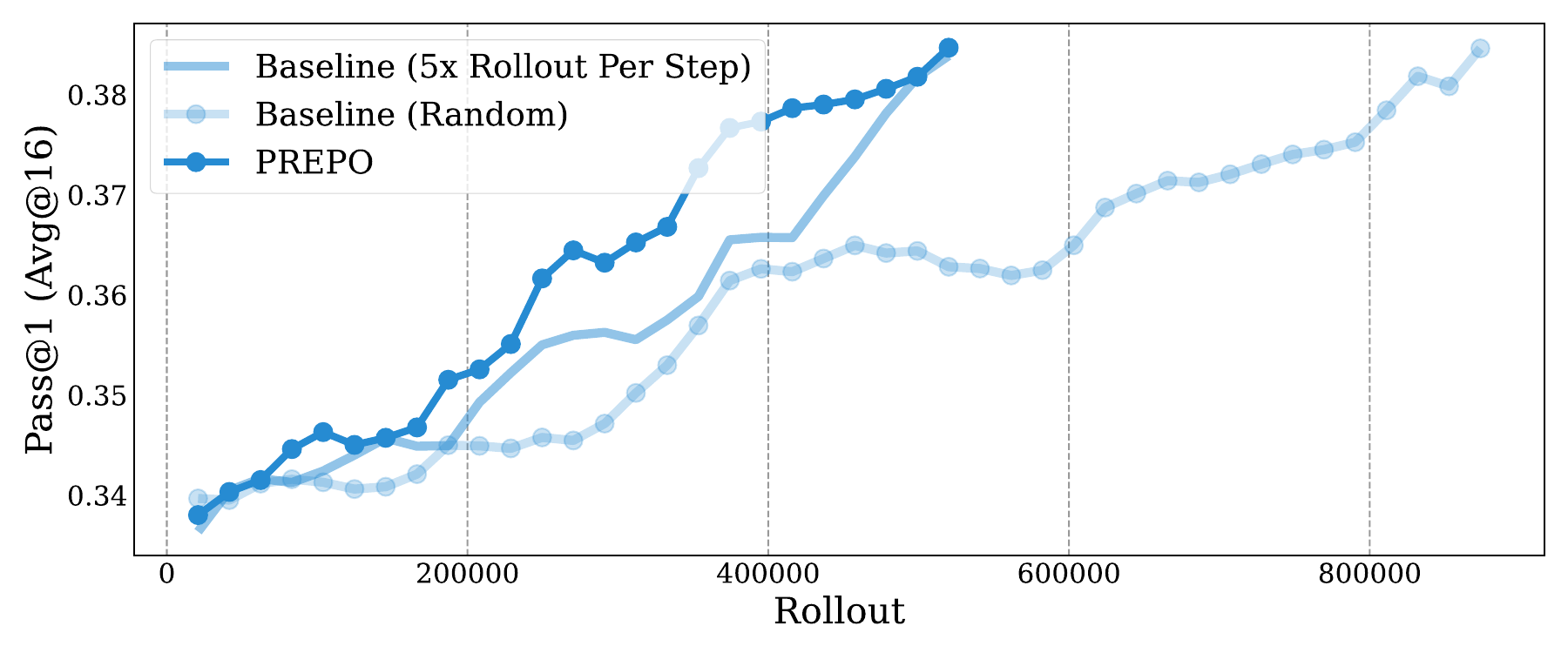}
    \caption{Comparison of PREPO and baselines (a) training w/o filtering (b) random selection.}
    \label{fig:full_data}
\end{figure}

For Qwen2.5-Math-7B, we observe that PREPO attains performance comparable to training on the full dataset without any filtering, i.e., using 5× rollouts per step, as shown in Figure \ref{fig:full_data} 



\section{Properties of PREPO}\label{app:theory}

\textbf{Setup.}  
The PREPO objective is defined as
\begin{equation}
\mathcal{J}_{\text{PREPO}}(\theta)
= \mathbb{E}_{\,x \sim \mathcal{I}_\rho,\;\{o_i\}_{i=1}^G \sim \pi_{\text{old}}}
\left[
\frac{1}{G}\sum_{i=1}^G w_i \cdot \frac{1}{|o_i|}\sum_{t=1}^{|o_i|}
s_{i,t}(\theta)\,\hat{A}_{i,t}
\right],
\end{equation}
where 
\[
s_{i,t}(\theta) = \frac{\pi_\theta(o_{i,t}\mid x,o_{i,<t})}{\pi_{\text{old}}(o_{i,t}\mid x,o_{i,<t})}, 
\qquad
\hat{A}_{i,t} = \frac{r_i - \operatorname{mean}(\{r_j\}_{j=1}^G)}{\operatorname{std}(\{r_j\}_{j=1}^G)}.
\]
Each rollout $i$ is further scaled by a sequence-level entropy weight
\[
\bar H_i = \frac{1}{|o_i|}\sum_{t=1}^{|o_i|} H_{i,t}, 
\qquad 
\bar H = \frac{1}{\sum_{k=1}^{B}|o_k|}\sum_{k=1}^{B}\sum_{t=1}^{|o_k|} H_{k,t}, 
\qquad 
w_i = \frac{\bar H_i}{\bar H},
\]
where $B$ denotes the micro-batch size used for a single gradient update.

\subsection{Sum of weights vs. batch size}  
The weights are normalized relative to $\bar H$, so that
\begin{equation}
\frac{1}{B}\sum_{i=1}^{B} w_i \cdot |o_i|
= \frac{1}{B\bar H}\sum_{i=1}^{B}|o_i|\bar H_i
= \frac{1}{B\bar H}\sum_{i=1}^{B}\sum_{t=1}^{|o_i|} H_{i,t}
= \frac{\sum_{i=1}^B |o_i|}{B}.
\end{equation}
Thus the \emph{token-weighted average weight equals the average sequence length}.  
If all sequences have equal length, then $\tfrac{1}{B}\sum_i w_i = 1$.  

\begin{figure}[H]
    \centering
    \includegraphics[width=0.5\linewidth]{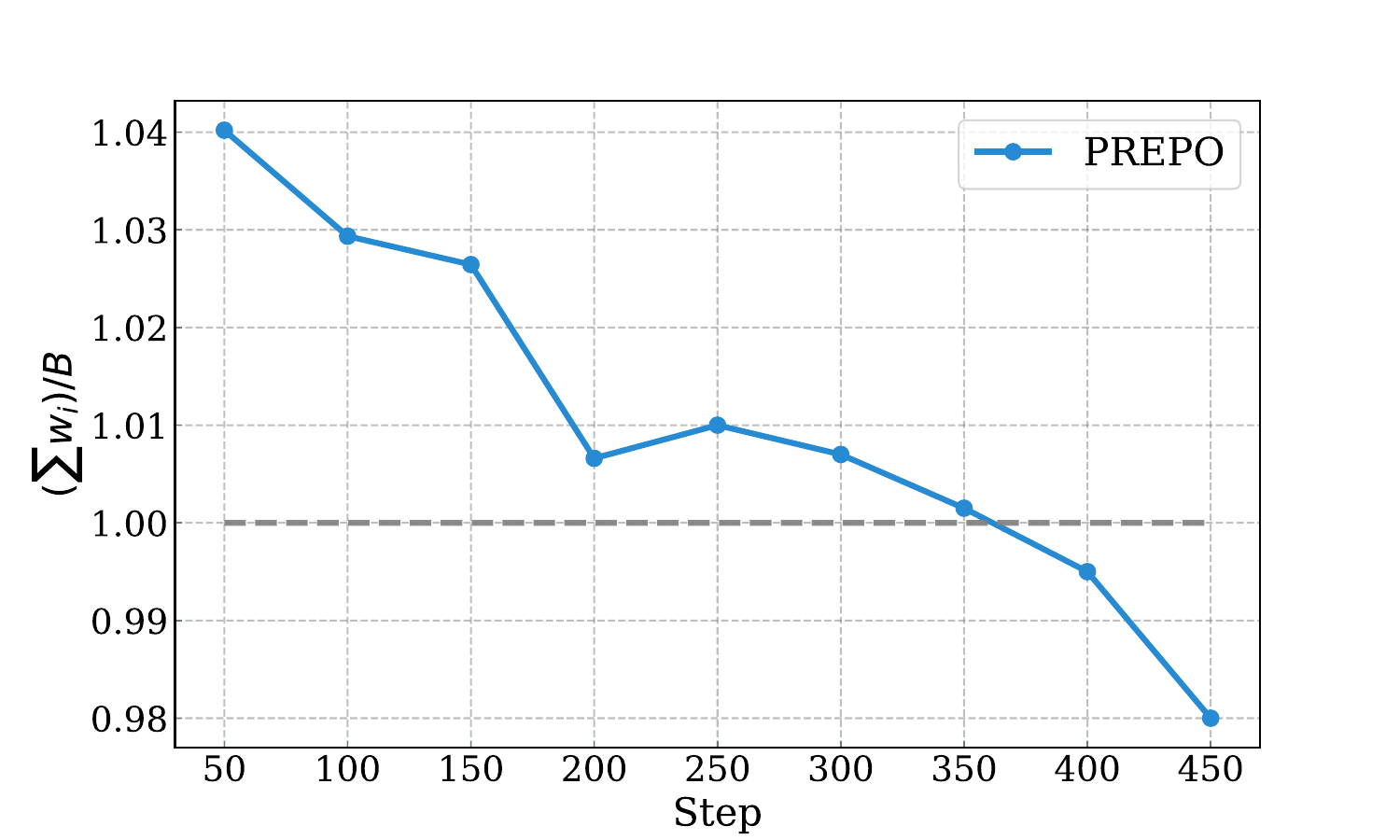}
    \caption{Trend of the effective batch size (Qwen2.5-Math-1.5B).}
    \label{fig:weight_trend}
\end{figure}

As shown in Figure~\ref{fig:weight_trend}, the effective batch size remains close to the nominal batch size throughout training. In our experiments, it starts at 1.04 and gradually declines to 0.98 during RLVR training. Early in training, when low-PPL prompts dominate and produce confident (low-entropy) responses, 
the average effective weight exceeds 1, reflecting a relative emphasis on the few higher-entropy rollouts. As training progresses and higher-PPL prompts enter, the overall entropy distribution shifts upward. After normalization, this causes the average effective weight to fall slightly below 1.

\subsection{Sensitivity to extreme entropies}  
Because $w_i=\bar H_i/\bar H$ is normalized by the batch mean, a rollout with extremely large entropy $\bar H_j \gg \bar H$ does not inflate its own weight ($w_j \!\approx\! 1$), but instead suppresses the weights of other rollouts ($w_i \!\ll\! 1$ for $i\ne j$). The partial derivative of the relative-entropy weight with respect to a sequence-level entropy in the batch is
\[
\frac{\partial w_i}{\partial \bar H_j}
= \begin{cases}
\dfrac{1}{\bar H} - \dfrac{\bar H_j}{\bar H^2}\dfrac{|o_j|}{\sum_k |o_k|}, & i=j, \\[2ex]
-\,\dfrac{\bar H_i}{\bar H^2}\dfrac{|o_j|}{\sum_k |o_k|}, & i\ne j,
\end{cases}
\]
which shows that (1) for the self-sensitivity ($i=j$), the two terms nearly cancel, so increasing $\bar H_j$ has only a small net effect on $w_j$ itself; (2) for cross-sensitivity ($i\neq j$), the derivative is strictly negative, meaning that increasing $\bar H_j$ decreases the weights of all other rollouts.

Thus, extreme entropies influence the distribution of weights not by amplifying the outlier’s own contribution, but by enlarging the batch mean $\bar H$, which in turn uniformly shrinks the normalized weights of the remaining rollouts in proportion to their entropies.

\begin{figure}[!h]
    \centering
    \includegraphics[width=0.5\linewidth]{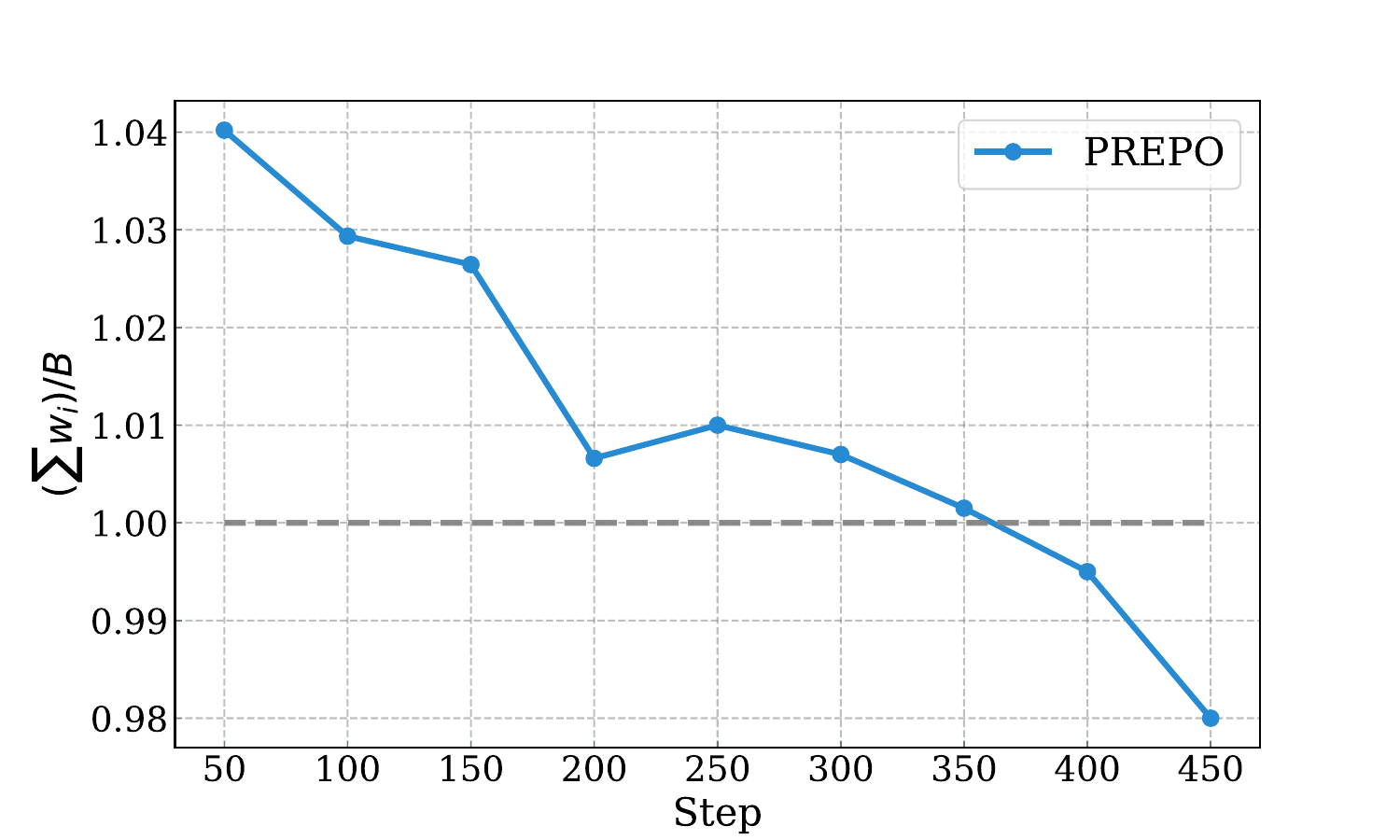}
    \caption{Frequency of relative-entropy weights (Qwen2.5-Math-1.5B; 50 training step)}
    \label{fig:weight_hist}
\end{figure}

As shown in Figure~\ref{fig:weight_hist}, the distribution of relative-entropy weights is sharply concentrated around 1, with most values between 0.7 and 1.5. A small fraction of rollouts appear in the long tail beyond 2, with rare outliers above 4. This pattern is consistent with the analysis above, that is, entropy outliers remain bounded in their own weights but shift the normalization, thereby downscaling the majority of rollouts.

\textbf{Remark on clipping.}  
For clarity, clipping was omitted from the above derivation. In practice, PPO-style clipping only truncates extreme importance ratios $s_{i,t}(\theta)$, effectively setting the corresponding gradient contributions to zero.





\newpage
\section*{NeurIPS Paper Checklist}

\begin{enumerate}

\item {\bf Claims}
    \item[] Question: Do the main claims made in the abstract and introduction accurately reflect the paper's contributions and scope?
    \item[] Answer: \answerYes{} 
    \item[] Justification: Our main claims are summarized in Figure \ref{fig:main} and Section \ref{para:experiments}.
    \item[] Guidelines:
    \begin{itemize}
        \item The answer NA means that the abstract and introduction do not include the claims made in the paper.
        \item The abstract and/or introduction should clearly state the claims made, including the contributions made in the paper and important assumptions and limitations. A No or NA answer to this question will not be perceived well by the reviewers. 
        \item The claims made should match theoretical and experimental results, and reflect how much the results can be expected to generalize to other settings. 
        \item It is fine to include aspirational goals as motivation as long as it is clear that these goals are not attained by the paper. 
    \end{itemize}

\item {\bf Limitations}
    \item[] Question: Does the paper discuss the limitations of the work performed by the authors?
    \item[] Answer: \answerYes{} 
    \item[] Justification: We include the limitations of our work in Section \ref{app:limitation}
    \item[] Guidelines:
    \begin{itemize}
        \item The answer NA means that the paper has no limitation while the answer No means that the paper has limitations, but those are not discussed in the paper. 
        \item The authors are encouraged to create a separate "Limitations" section in their paper.
        \item The paper should point out any strong assumptions and how robust the results are to violations of these assumptions (e.g., independence assumptions, noiseless settings, model well-specification, asymptotic approximations only holding locally). The authors should reflect on how these assumptions might be violated in practice and what the implications would be.
        \item The authors should reflect on the scope of the claims made, e.g., if the approach was only tested on a few datasets or with a few runs. In general, empirical results often depend on implicit assumptions, which should be articulated.
        \item The authors should reflect on the factors that influence the performance of the approach. For example, a facial recognition algorithm may perform poorly when image resolution is low or images are taken in low lighting. Or a speech-to-text system might not be used reliably to provide closed captions for online lectures because it fails to handle technical jargon.
        \item The authors should discuss the computational efficiency of the proposed algorithms and how they scale with dataset size.
        \item If applicable, the authors should discuss possible limitations of their approach to address problems of privacy and fairness.
        \item While the authors might fear that complete honesty about limitations might be used by reviewers as grounds for rejection, a worse outcome might be that reviewers discover limitations that aren't acknowledged in the paper. The authors should use their best judgment and recognize that individual actions in favor of transparency play an important role in developing norms that preserve the integrity of the community. Reviewers will be specifically instructed to not penalize honesty concerning limitations.
    \end{itemize}

\item {\bf Theory assumptions and proofs}
    \item[] Question: For each theoretical result, does the paper provide the full set of assumptions and a complete (and correct) proof?
    \item[] Answer: \answerNA{} 
    \item[] Justification: This is not a theoretical paper.
    \item[] Guidelines:
    \begin{itemize}
        \item The answer NA means that the paper does not include theoretical results. 
        \item All the theorems, formulas, and proofs in the paper should be numbered and cross-referenced.
        \item All assumptions should be clearly stated or referenced in the statement of any theorems.
        \item The proofs can either appear in the main paper or the supplemental material, but if they appear in the supplemental material, the authors are encouraged to provide a short proof sketch to provide intuition. 
        \item Inversely, any informal proof provided in the core of the paper should be complemented by formal proofs provided in appendix or supplemental material.
        \item Theorems and Lemmas that the proof relies upon should be properly referenced. 
    \end{itemize}

    \item {\bf Experimental result reproducibility}
    \item[] Question: Does the paper fully disclose all the information needed to reproduce the main experimental results of the paper to the extent that it affects the main claims and/or conclusions of the paper (regardless of whether the code and data are provided or not)?
    \item[] Answer: \answerYes{} 
    \item[] Justification: We explained our settings in Section \ref{para:experiments}.
    \item[] Guidelines: 
    \begin{itemize}
        \item The answer NA means that the paper does not include experiments.
        \item If the paper includes experiments, a No answer to this question will not be perceived well by the reviewers: Making the paper reproducible is important, regardless of whether the code and data are provided or not.
        \item If the contribution is a dataset and/or model, the authors should describe the steps taken to make their results reproducible or verifiable. 
        \item Depending on the contribution, reproducibility can be accomplished in various ways. For example, if the contribution is a novel architecture, describing the architecture fully might suffice, or if the contribution is a specific model and empirical evaluation, it may be necessary to either make it possible for others to replicate the model with the same dataset, or provide access to the model. In general. releasing code and data is often one good way to accomplish this, but reproducibility can also be provided via detailed instructions for how to replicate the results, access to a hosted model (e.g., in the case of a large language model), releasing of a model checkpoint, or other means that are appropriate to the research performed.
        \item While NeurIPS does not require releasing code, the conference does require all submissions to provide some reasonable avenue for reproducibility, which may depend on the nature of the contribution. For example
        \begin{enumerate}
            \item If the contribution is primarily a new algorithm, the paper should make it clear how to reproduce that algorithm.
            \item If the contribution is primarily a new model architecture, the paper should describe the architecture clearly and fully.
            \item If the contribution is a new model (e.g., a large language model), then there should either be a way to access this model for reproducing the results or a way to reproduce the model (e.g., with an open-source dataset or instructions for how to construct the dataset).
            \item We recognize that reproducibility may be tricky in some cases, in which case authors are welcome to describe the particular way they provide for reproducibility. In the case of closed-source models, it may be that access to the model is limited in some way (e.g., to registered users), but it should be possible for other researchers to have some path to reproducing or verifying the results.
        \end{enumerate}
    \end{itemize}

\item {\bf Open access to data and code}
    \item[] Question: Does the paper provide open access to the data and code, with sufficient instructions to faithfully reproduce the main experimental results, as described in supplemental material?
    \item[] Answer: \answerYes{} 
    \item[] Justification: Github link provided.
    \item[] Guidelines: 
    \begin{itemize}
        \item The answer NA means that paper does not include experiments requiring code.
        \item Please see the NeurIPS code and data submission guidelines (\url{https://nips.cc/public/guides/CodeSubmissionPolicy}) for more details.
        \item While we encourage the release of code and data, we understand that this might not be possible, so “No” is an acceptable answer. Papers cannot be rejected simply for not including code, unless this is central to the contribution (e.g., for a new open-source benchmark).
        \item The instructions should contain the exact command and environment needed to run to reproduce the results. See the NeurIPS code and data submission guidelines (\url{https://nips.cc/public/guides/CodeSubmissionPolicy}) for more details.
        \item The authors should provide instructions on data access and preparation, including how to access the raw data, preprocessed data, intermediate data, and generated data, etc.
        \item The authors should provide scripts to reproduce all experimental results for the new proposed method and baselines. If only a subset of experiments are reproducible, they should state which ones are omitted from the script and why.
        \item At submission time, to preserve anonymity, the authors should release anonymized versions (if applicable).
        \item Providing as much information as possible in supplemental material (appended to the paper) is recommended, but including URLs to data and code is permitted.
    \end{itemize}

\item {\bf Experimental setting/details}
    \item[] Question: Does the paper specify all the training and test details (e.g., data splits, hyperparameters, how they were chosen, type of optimizer, etc.) necessary to understand the results?
    \item[] Answer: \answerYes{} 
    \item[] Justification: Details are summarized in Section \ref{para:experiments}.
    \item[] Guidelines:
    \begin{itemize}
        \item The answer NA means that the paper does not include experiments.
        \item The experimental setting should be presented in the core of the paper to a level of detail that is necessary to appreciate the results and make sense of them.
        \item The full details can be provided either with the code, in appendix, or as supplemental material.
    \end{itemize}

\item {\bf Experiment statistical significance}
    \item[] Question: Does the paper report error bars suitably and correctly defined or other appropriate information about the statistical significance of the experiments?
    \item[] Answer: \answerYes{} 
    \item[] Justification: We show the average performance of over multiple runs.
    \item[] Guidelines:
    \begin{itemize}
        \item The answer NA means that the paper does not include experiments.
        \item The authors should answer "Yes" if the results are accompanied by error bars, confidence intervals, or statistical significance tests, at least for the experiments that support the main claims of the paper.
        \item The factors of variability that the error bars are capturing should be clearly stated (for example, train/test split, initialization, random drawing of some parameter, or overall run with given experimental conditions).
        \item The method for calculating the error bars should be explained (closed form formula, call to a library function, bootstrap, etc.)
        \item The assumptions made should be given (e.g., Normally distributed errors).
        \item It should be clear whether the error bar is the standard deviation or the standard error of the mean.
        \item It is OK to report 1-sigma error bars, but one should state it. The authors should preferably report a 2-sigma error bar than state that they have a 96\% CI, if the hypothesis of Normality of errors is not verified.
        \item For asymmetric distributions, the authors should be careful not to show in tables or figures symmetric error bars that would yield results that are out of range (e.g. negative error rates).
        \item If error bars are reported in tables or plots, The authors should explain in the text how they were calculated and reference the corresponding figures or tables in the text.
    \end{itemize}

\item {\bf Experiments compute resources}
    \item[] Question: For each experiment, does the paper provide sufficient information on the computer resources (type of compute workers, memory, time of execution) needed to reproduce the experiments?
    \item[] Answer: \answerYes{} 
    \item[] Justification: Details are summarized in Section \ref{para:experiments}.
    \item[] Guidelines: 
    \begin{itemize}
        \item The answer NA means that the paper does not include experiments.
        \item The paper should indicate the type of compute workers CPU or GPU, internal cluster, or cloud provider, including relevant memory and storage.
        \item The paper should provide the amount of compute required for each of the individual experimental runs as well as estimate the total compute. 
        \item The paper should disclose whether the full research project required more compute than the experiments reported in the paper (e.g., preliminary or failed experiments that didn't make it into the paper). 
    \end{itemize}
    
\item {\bf Code of ethics}
    \item[] Question: Does the research conducted in the paper conform, in every respect, with the NeurIPS Code of Ethics \url{https://neurips.cc/public/EthicsGuidelines}?
    \item[] Answer: \answerYes{} 
    \item[] Justification: We have read the understood the code of ethics; and have done our best to conform.
    \item[] Guidelines:
    \begin{itemize}
        \item The answer NA means that the authors have not reviewed the NeurIPS Code of Ethics.
        \item If the authors answer No, they should explain the special circumstances that require a deviation from the Code of Ethics.
        \item The authors should make sure to preserve anonymity (e.g., if there is a special consideration due to laws or regulations in their jurisdiction).
    \end{itemize}

\item {\bf Broader impacts}
    \item[] Question: Does the paper discuss both potential positive societal impacts and negative societal impacts of the work performed?
    \item[] Answer: \answerNA{} 
    \item[] Justification: Our work does not impact the society at large.
    \item[] Guidelines:
    \begin{itemize}
        \item The answer NA means that there is no societal impact of the work performed.
        \item If the authors answer NA or No, they should explain why their work has no societal impact or why the paper does not address societal impact.
        \item Examples of negative societal impacts include potential malicious or unintended uses (e.g., disinformation, generating fake profiles, surveillance), fairness considerations (e.g., deployment of technologies that could make decisions that unfairly impact specific groups), privacy considerations, and security considerations.
        \item The conference expects that many papers will be foundational research and not tied to particular applications, let alone deployments. However, if there is a direct path to any negative applications, the authors should point it out. For example, it is legitimate to point out that an improvement in the quality of generative models could be used to generate deepfakes for disinformation. On the other hand, it is not needed to point out that a generic algorithm for optimizing neural networks could enable people to train models that generate Deepfakes faster.
        \item The authors should consider possible harms that could arise when the technology is being used as intended and functioning correctly, harms that could arise when the technology is being used as intended but gives incorrect results, and harms following from (intentional or unintentional) misuse of the technology.
        \item If there are negative societal impacts, the authors could also discuss possible mitigation strategies (e.g., gated release of models, providing defenses in addition to attacks, mechanisms for monitoring misuse, mechanisms to monitor how a system learns from feedback over time, improving the efficiency and accessibility of ML).
    \end{itemize}
    
\item {\bf Safeguards}
    \item[] Question: Does the paper describe safeguards that have been put in place for responsible release of data or models that have a high risk for misuse (e.g., pretrained language models, image generators, or scraped datasets)?
    \item[] Answer: \answerNA{} 
    \item[] Justification: Since we use open-source models and datasets, our work poses no risk of misuse.
    \item[] Guidelines:
    \begin{itemize}
        \item The answer NA means that the paper poses no such risks.
        \item Released models that have a high risk for misuse or dual-use should be released with necessary safeguards to allow for controlled use of the model, for example by requiring that users adhere to usage guidelines or restrictions to access the model or implementing safety filters. 
        \item Datasets that have been scraped from the Internet could pose safety risks. The authors should describe how they avoided releasing unsafe images.
        \item We recognize that providing effective safeguards is challenging, and many papers do not require this, but we encourage authors to take this into account and make a best faith effort.
    \end{itemize}

\item {\bf Licenses for existing assets}
    \item[] Question: Are the creators or original owners of assets (e.g., code, data, models), used in the paper, properly credited and are the license and terms of use explicitly mentioned and properly respected?
    \item[] Answer: \answerYes{} 
    \item[] Justification: We cited open-sourced libraries in our paper.
    \item[] Guidelines:
    \begin{itemize}
        \item The answer NA means that the paper does not use existing assets.
        \item The authors should cite the original paper that produced the code package or dataset.
        \item The authors should state which version of the asset is used and, if possible, include a URL.
        \item The name of the license (e.g., CC-BY 4.0) should be included for each asset.
        \item For scraped data from a particular source (e.g., website), the copyright and terms of service of that source should be provided.
        \item If assets are released, the license, copyright information, and terms of use in the package should be provided. For popular datasets, \url{paperswithcode.com/datasets} has curated licenses for some datasets. Their licensing guide can help determine the license of a dataset.
        \item For existing datasets that are re-packaged, both the original license and the license of the derived asset (if it has changed) should be provided.
        \item If this information is not available online, the authors are encouraged to reach out to the asset's creators.
    \end{itemize}

\item {\bf New assets}
    \item[] Question: Are new assets introduced in the paper well documented and is the documentation provided alongside the assets?
    \item[] Answer: \answerYes{} 
    \item[] Justification: We will release our code base with included readme files.
    \item[] Guidelines:
    \begin{itemize}
        \item The answer NA means that the paper does not release new assets.
        \item Researchers should communicate the details of the dataset/code/model as part of their submissions via structured templates. This includes details about training, license, limitations, etc. 
        \item The paper should discuss whether and how consent was obtained from people whose asset is used.
        \item At submission time, remember to anonymize your assets (if applicable). You can either create an anonymized URL or include an anonymized zip file.
    \end{itemize}

\item {\bf Crowdsourcing and research with human subjects}
    \item[] Question: For crowdsourcing experiments and research with human subjects, does the paper include the full text of instructions given to participants and screenshots, if applicable, as well as details about compensation (if any)? 
    \item[] Answer: \answerNA{} 
    \item[] Guidelines: This work does not involve crowdsourcing nor research within human subjects.
    \begin{itemize}
        \item The answer NA means that the paper does not involve crowdsourcing nor research with human subjects.
        \item Including this information in the supplemental material is fine, but if the main contribution of the paper involves human subjects, then as much detail as possible should be included in the main paper. 
        \item According to the NeurIPS Code of Ethics, workers involved in data collection, curation, or other labor should be paid at least the minimum wage in the country of the data collector. 
    \end{itemize}

\item {\bf Institutional review board (IRB) approvals or equivalent for research with human subjects}
    \item[] Question: Does the paper describe potential risks incurred by study participants, whether such risks were disclosed to the subjects, and whether Institutional Review Board (IRB) approvals (or an equivalent approval/review based on the requirements of your country or institution) were obtained?
    \item[] Answer: \answerNA{} 
    \item[] Justification: This work does not involve crowdsourcing nor research within human subjects.
    \item[] Guidelines:
    \begin{itemize}
        \item The answer NA means that the paper does not involve crowdsourcing nor research with human subjects.
        \item Depending on the country in which research is conducted, IRB approval (or equivalent) may be required for any human subjects research. If you obtained IRB approval, you should clearly state this in the paper. 
        \item We recognize that the procedures for this may vary significantly between institutions and locations, and we expect authors to adhere to the NeurIPS Code of Ethics and the guidelines for their institution. 
        \item For initial submissions, do not include any information that would break anonymity (if applicable), such as the institution conducting the review.
    \end{itemize}

\item {\bf Declaration of LLM usage}
    \item[] Question: Does the paper describe the usage of LLMs if it is an important, original, or non-standard component of the core methods in this research? Note that if the LLM is used only for writing, editing, or formatting purposes and does not impact the core methodology, scientific rigorousness, or originality of the research, declaration is not required.
    \item[] Answer: \answerYes{} 
    \item[] Justification: In this work, LLMs were employed
solely for grammar correction and sentence rephrasing. They had no involvement in research ideation,
experimental design, data analysis, or substantive writing. Their role was restricted to improving
clarity and style; therefore, they are not considered contributors to the research. 
    \item[] Guidelines:
    \begin{itemize}
        \item The answer NA means that the core method development in this research does not involve LLMs as any important, original, or non-standard components.
        \item Please refer to our LLM policy (\url{https://neurips.cc/Conferences/2025/LLM}) for what should or should not be described.
    \end{itemize}

\end{enumerate}

\end{document}